\documentclass[journal]{IEEEtran} 
\newcommand{\etal}{\emph{et al.}}
\usepackage{times}
\let\labelindent\relax

\usepackage[numbers,sort&compress]{natbib}

\usepackage{multicol}
\usepackage[bookmarks=true]{hyperref}
\usepackage{tikz}
\usetikzlibrary{bayesnet}
\IEEEoverridecommandlockouts
\usepackage{graphicx}

\usepackage{subfig} % For subfloat
\usepackage[skip=0pt,font=small,labelfont=bf]{caption} % Remove spacing between figs and captions
\usepackage{textcomp}
\usepackage{color}
\usepackage{mathtools}
\usepackage{amsmath}
\usepackage{enumitem}
\usepackage{flushend}

\usepackage{epsfig,graphicx,amsmath,amsfonts}
\usepackage{import}
\usepackage{transparent}
\usepackage{upgreek}
\usepackage[export]{adjustbox}

\DeclareMathOperator{\Tr}{Tr}

\newcommand\numberthis{\addtocounter{equation}{1}\tag{\theequation}}

\newcommand{\W}[1]{\prescript{w}{}{#1}}
\newcommand{\B}[1]{\prescript{b}{}{#1}}
\newcommand{\T}[1]{\prescript{t-2}{}{#1}}

\usepackage[normalem]{ulem}
\usepackage{multirow}
\usepackage{float}
\usepackage[utf8]{inputenc}
\usepackage[TS1,T1]{fontenc}
\usepackage{array, booktabs}
\usepackage{xcolor}

\usepackage{graphicx}
\usepackage{comment}
\usepackage{colortbl}
\usepackage{caption}
\usepackage[color=green!20]{todonotes}

\definecolor{gt_color}{HTML}{FFFFA1}
\definecolor{bg_color}{HTML}{FFFFFF}
\definecolor{bt_color}{HTML}{C0DFA1}
\definecolor{rev_color}{rgb}{1.0,1.0,0}

\usepackage{color,soul}
\sethlcolor{rev_color}
\newcommand{\HL}[1]{#1}

\title{\LARGE \bf In-Hand Object-Dynamics Inference using Tactile Fingertips}
\author{Balakumar Sundaralingam$^{1,2,*}$ and Tucker Hermans$^{1,2}$ %
  \thanks{$^1$~University of Utah Robotics Center and the School of Computing, University of Utah, Salt Lake City, UT, USA. Email: {\tt\small\{bala, thermans\}@cs.utah.edu}}
  \thanks{$^2$ NVIDIA, USA}
\thanks{$^*$ work done while B. Sundaralingam was affiliated to~$^1$.}}

\begin{document}
%\iffalse
\setcounter{figure}{1}
\makeatletter
\let\@oldmaketitle\@maketitle% Store \@maketitle
\renewcommand{\@maketitle}{\@oldmaketitle% Update \@maketitle to insert...
\begin{center}
    \centering     
    \includegraphics[width=\textwidth]
    {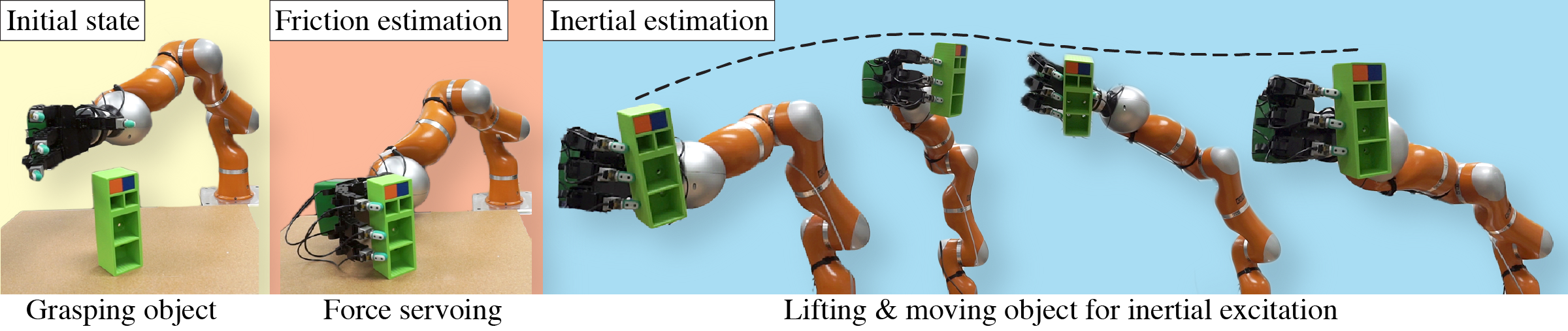}
  \end{center}
  %\refstepcounter{figure}
  \footnotesize{\textbf{Fig.~\thefigure:\label{fig:intro}}~We propose in-hand object dynamics estimation leveraging fingertip tactile force perception. This enables a robot to make contact with the object, perform friction estimation while being supported by the surface. Inertial estimation can then be performed by exciting the object dynamics in-hand, by lifting the object and moving along trajectories.}\vspace{-12pt}
  \medskip}% ... an image
\makeatother
%\fi
\maketitle
\begin{abstract}
  Having the ability to estimate an object's properties through interaction will enable robots to manipulate novel objects. Object's dynamics, specifically the friction and inertial parameters have only been estimated in a lab environment with precise and often external sensing. Could we infer an object's dynamics in the wild with only the robot's sensors?  In this paper, we explore the estimation of dynamics of a grasped object in motion, with tactile force sensing at multiple fingertips. Our estimation approach does not rely on torque sensing to estimate the dynamics. To estimate friction, we develop a control scheme to actively interact with the object until slip is detected. To robustly perform the inertial estimation, we setup a factor graph that fuses all our sensor measurements on physically consistent manifolds and perform inference.  We show that tactile fingertips enable in-hand dynamics estimation of low mass objects.
\end{abstract}

\section{Introduction}
\label{sec:introduction}
Acquiring the physical properties of previously unencountered objects could enable robots to robustly interact in the unstructured real world~\cite{tsikos1991segmentation, salganicoff1991sensorimotor, salganicoff1994active, katz2008manipulating, martin2014online, royadynamic,mavrakis2020estimation}. While vision has enabled object geometry completion from partial views~\cite{park2019deepsdf,merwe2019learning}, knowledge about an object's dynamics is hard to obtain from viewing a static state. Leveraging the interaction capabilities in robotic systems can make object dynamics estimation possible from observing sensor measurements. Motivated by recent studies on interactive perception~\cite{bohg2017interactive}, we explore estimating object dynamics, specifically the friction, and inertial parameters, of an object with known geometry using vision and tactile fingertip sensing as shown in Fig.~1.

Given knowledge of object dynamics, robots can perform accurate and robust multi-contact manipulation of modeled objects via physics-based planning and control ~\cite{an_dynmodel_control_1989,ferrari1992planning,erdmann1993multiple,Han1998,Mordatch2012a,chavan2018hand,shi2019inhand,mavrakis2020survey}. Physically realistic simulation could also benefit from real-world object dynamics knowledge to train policies in simulation that robustly transfer to real systems. For example, estimates of parameters' distribution could serve as bounds for domain randomization~\cite{chebotar2019closing}. Object dynamics knowledge also aids in improved perception of robot object interaction even under noisy sensor observations compared to only relying on geometry~\cite{lambert-2019icra-vistac}. 

Previously, object inertial estimation has been studied to improve industrial manipulator performance~\cite{Atkeson1986} by attaching the object of interest as a rigid load to a wrist-mounted \HL{force torque} sensor. By tracking the object acceleration and the \HL{force torque} readings, one can estimate the inertial parameters in a least squares formulation~\cite{Atkeson1986}. However, for robots with dexterous hands, a wrist-mounted \HL{force torque} sensor cannot directly be used for estimation as: 1)~\HL{Force torque} sensors require calibration and cannot obtain sufficient accuracy for object inertial estimation with in-situ calibration~\cite{chavez2019model}, 2)~Negating the \HL{force torque} readings due to the hand from the sensed wrist-mounted \HL{force torque} values is noise prone as most dexterous hands have noisy or non-existent joint state observations. These issues become more apparent when the object's inertial properties are significantly smaller than the hand's inertial properties.

To overcome these limitations in using wrist-mounted \HL{force torque} sensors for grasped object dynamics estimation, we leverage tactile sensors available on the fingertips of common dexterous hands~\cite{dahiya2010tactile, kappassov2015tactile, johnston1996full, wettels2014multimodal}. Specifically, we show our proposed method working on one popular sensor--the BioTac, a biomimetic tactile sensor which can estimate physical quantities such as force~\cite{wettels2011haptic,Sundaralingam-ICRA-19} and point of contact~\cite{loeb2013estimating}. Estimating object dynamics also requires accurate measurements of object acceleration. Researchers have relied on motion capture systems to obtain object acceleration~\cite{kolev2015physically}, however this is not ideal for the unstructured world. Hence, we explore leveraging a vision-based markerless object tracker~\cite{schmidt2015depth} and an in-grasp kinematic model~\cite{sundaralingam2019relaxed} based object tracker to enable observing object acceleration.

To enable object dynamics estimation under noisy observations, we encode the dynamics existent between the fingertips and the grasped object in a factor graph with the robot's sensor observations. Our factor graph contains physically consistent projections and euclidean retractions of the inertial parameters to ensure physically correct step update of the parameters using standard numerical optimization solvers. Our structured approach to inertial estimation enables optimizing over multiple costs on different manifolds as we retract all the parameters to the euclidean space before updating the parameters. We also directly encode object pose observations and compute object accelerations through finite differencing in the factor graph.

Apart from inertial parameters, friction also plays an important role in contact-based object interaction such as in grasp planning via force-closure optimization~\cite{miller2004graspit} and sliding based in-hand manipulation~\cite{shi2019inhand}. Empirical studies of planar manipulation have also shown friction to play a key role in how objects react to contact~\cite{chavan2018hand,bauza18a}. In this paper, we explore an in-hand motion generation scheme for estimating the static coefficient of friction based on detection of slip as a first step towards accurate in-hand object friction estimation.

We summarize our contributions, validated with real-world experiments below,

\noindent \textbf{In-Hand Friction Cone Inference:} A force-slip motion generation scheme that servoes a fingertip on the object while other fingertips maintain a rigid hold on the object to infer the coefficient of static friction.

\noindent \textbf{SYS-ID from Force Fingertips:} We formulate the Newton-Euler equations to estimate inertial parameters from force sensing at fingertips which is novel from existing methods that rely on torque sensing to estimate inertial parameters.

\noindent \textbf{Inertial Estimation as Structured Inference:} We formulate the object dynamics inference problem as inference in a factor graph, with physically consistency encoded by manifolds.

\noindent \textbf{Multi-Contact Dataset:} We are releasing a dataset that contains a robot interacting with a 3D printed object containing over 200k time samples and also share CAD model of the object. \HL{This is available at}~\url{https://sites.google.com/view/tactile-obj-dynamics}

We conduct a thorough review of literature in Sec.~\ref{sec:related-work}. A formal definition of the problem and the proposed approach is discussed next in Sec.~\ref{sec:prob-app}, followed by implementation details and evaluation metrics in Sec.~\ref{sec:implement}. We report the estimation results in Sec.~\ref{sec:results-discussion} followed by a discussion in Sec.~\ref{sec:discussion}. We then leave concluding remarks in Sec.~\ref{sec:conclusion}.

%%% Local Variables:
%%% mode: latex
%%% TeX-master: "../paperdraft"
%%% End:

\section{Related Work}
\label{sec:related-work}
In this literature review, we focus on methods that infer object dynamics in the physical parametric form, specifically the inertial and friction parameters. We discuss their contributions and highlight the novelty in our approach.  We will first discuss object dynamics estimation methods broadly across robotics~\cite{yoshida1996set,Atkeson1986,li1989indirect,Franchi2014,Yu1999estimation,Yu2005,traversaro2016identification,wensing2018linear,lee2018geometric,Luo2015,murooka2018simultaneous,murotsu1994parameter,ma2008orbit,Fazeli2017}, followed by methods that estimate object properties through fingertip sensing~\cite{Yu1999estimation,Yu2005,Zhao2018,Franchi2014,Liu2011,luo2017robotic,Song2014,trkov2015stick,khamis2018papillarray,xiang2019stick,su2015force,nakamura2001tactile,maeno2004friction,okatani2016tactile2,herrera2008multilayer}.

When estimating the inertial matrix, it is essential for the matrix to be physically consistent as shown by Yoshida~\etal~\cite{yoshida1996set}. Atkeson~\etal~\cite{Atkeson1986} rely on post processing to ensure physical consistency while Li and Slotine~\cite{li1989indirect} examine incorporating physical consistency during online estimation for adaptive control. Recent methods explore incorporating physically consistency as part of the estimation to enable better recovery of parameters~\cite{traversaro2016identification,wensing2018linear,lee2018geometric}.

Traversaro~\etal~\cite{traversaro2016identification} formulate inertial estimation as optimization over the \HL{3D rotation group}~($\mathbf{SO}(3)$) manifold in addition to other linear inertial parameters and leverage a custom manifold optimization solver to estimate physically consistent inertial parameters. Wensing~\etal~\cite{wensing2018linear} extend these manifold constraints to convex constraints with the use of linear matrix inequalities~(LMI). Though the LMIs enable physical consistency of the estimated parameters, the parameters might still exist near the boundary of the constraints~\cite{lee2018geometric}. Lee and Park~\cite{lee2018geometric} show the advantage of using Reimmannian metrics instead of the standard l2 norm for regularization of the inertial parameters for accurate parameter estimation. However, their method performs significantly slower at 23 minutes compared to 6.3 seconds by Wensing~\etal~\cite{wensing2018linear}. Nevertheless, Lee~\etal's~\cite{lee2018geometric} method estimate does not overfit to the data while Wensing~\etal
's~\cite{wensing2018linear} overfits to seen data. We combine the physical consistency ideas developed by~Traversaro~\etal~\cite{traversaro2016identification} and Lee and Park~\cite{lee2018geometric} in a single optimization framework to optimize over both the~$\mathbf{SO}(3)$ and inertial manifolds. 

A common assumption across methods is the use of accurate object pose estimates either from a marker based high speed vision system, joint angle measurements when the object is rigidly attached or using skeleton tracking. The data is pre-processed to a set of \HL{force torque} readings and accelerations. This process of performing estimation will not work when the noise in observing accelerations~(e.g., tracking small objects with a marker-less system) is large.  We do not make this assumption, rather solving the inertial estimation as inference in a factor graph with noisy observations of the object pose.

We now discuss methods that perform object dynamics estimation from fingertip sensing. Yu~\etal~\cite{Yu1999estimation} explore estimation of mass and center of mass of objects by tipping with a single force sensing fingertip. Their estimation works only if the contact line between the object and the support surface remains fixed during the multiple tipping operations, making it very hard to work with non-box shaped objects such as cylinders and spheres. Light weight objects are also hard to use as they can slide during tipping. Yu~\etal~\cite{Yu2005} explore estimation of inertial parameters through planar pushing with two  force fingertips but do not attempt friction estimation. They also can only estimate center of mass and inertia in \HL{the special Euclidean group in the plane~($\mathbf{SE}(2)$)}.

Zhao~\etal~\cite{Zhao2018} explore estimation of CoM and friction coefficient by detection of rotation slip at the tactile fingertips and a load cell which measures the force due to gravity on the object. They experimentally validate on simple shaped objects with a gripper. Their approach is formulated only for finding the center of mass along two axes.

\HL{Force torque} sensing from multiple locations has been studied in mobile robots by 
Franchi~\etal~\cite{Franchi2014}. They focus on inertial and kinematic parameter estimation of an unknown load being manipulated by a team of mobile robots. They explore inertial estimation through distrubuted estimation filters with control actions that ensure the observability of the parameters. Their approach however assumes the object is supported by wheels and do not estimate the mass or friction.

Murooka~\etal~\cite{murooka2018simultaneous} explore simultaneous planning and estimation for humanoid tasks in the real world. They assume the manipulation to be quasi-static and estimate center of mass and friction of objects with known mass, geometry and supported by the ground. They solve a quadratic program to estimate the ground contact force given the robot's contact force measurements. 

In contrast to methods that have leveraged sensing at fingertips/multiple contacts~\cite{Yu1999estimation,Yu2005,Zhao2018,Franchi2014,murooka2018simultaneous}, our approach leverages the linear forces~(i.e., no torques) estimated at the fingertips from tactile signals when the object is grasped in-hand to estimate the object dynamics. 
Researchers have also attempted to estimate inertial parameters without using force or torque measurements~\cite{murotsu1994parameter,ma2008orbit,Fazeli2017}. However these methods have very specific assumptions, e.g., Fazeli~\etal~\cite{Fazeli2017} has only been shown to work in~$\mathbf{SE}(2)$.

Friction forces prove a complex phenomenon to  model and estimate from sensing only object motion and forces as shown by~\cite{Lynch1993,sinha1993robotic,Luo2015}. Tactile sensing at contacts can enable capturing the rich friction information directly as shown by studies on human computer interaction systems~\cite{wiertlewski2016partial,dzidek2017pens,gueorguiev2017tactile}. Tactile sensing for friction estimation has been used in two primary ways--studying the deformation on the soft tactile skin during contact with an object~\cite{nakamura2001tactile,maeno2004friction,okatani2016tactile2,herrera2008multilayer} and detecting the tangential force at the point of slip~\cite{Song2014,trkov2015stick,khamis2018papillarray,xiang2019stick,su2015force}.

Methods that study the deformation of the soft skin primarily leverage detection of the rate of incipient slip either using multiple inertial measurement units~(IMUs) or through finite element methods~\cite{maeno2004friction}. This rate of change of incipient slip allows for estimating the coefficient of friction. However, this approach has only been shown to work with bulky and large tactile sensors. Methods that have used tangential force measurement before point of slip have either relied on gravity to cause slippage~\cite{su2015force} or require a rigid mount of the object to perform estimation~\cite{Song2014}.  At a broader scale, friction has also been indirectly estimated in terms of slip prediction from tactile signals~\cite{veiga-iros2015-slip-learning,yin2018measuring,fujimoto2003development} and surface classification~\cite{Liu2011,luo2017robotic}. These indirect estimation methods only enable task specific feedback and do not give a parametric value for friction.

In our approach to friction estimation, we actively generate motions for the fingertip on an object held in a grasp to find the boundaries of the friction cone. At this boundary, the fingertip would slip and the force before slippage is used to estimate the friction. Our approach enables estimating friction only relies on force estimation at the fingertips.

%%% Local Variables:
%%% mode: latex
%%% TeX-master: "../paperdraft"
%%% End:

\section{Problem Formulation \& Proposed Approach} 
\label{sec:prob-app}
Consider~$n$ tactile fingertips making contacts~\HL{$p_{i \in [1,n]}$} on an object as shown in Fig.~\ref{fig:notations}. Each tactile fingertip~$i$ \HL{senses} a force~$f_{i}$ between the object and the fingertip. We define the problem as estimating the dynamic properties of the object by leveraging sensing from these~$n$ tactile force fingertips.  The dynamic properties being the static friction coefficient~\HL{$\mu_s \in \mathbb{R}^+$}, the object's mass~\HL{$m \in \mathbb{R}^+$}, center of mass~$r_c \in \mathbb{R}^3$ and the inertia matrix~$H_{cm} \in \mathbb{R}^6$. We define the inertia matrix~$H_{cm}$ to consist of moments of inertia~$[H_{xx},H_{yy},H_{zz}]\in\mathbb{R}^3$ and the products of inertia~$[H_{xy},H_{xz},H_{yz}]\in \mathbb{R}^3$ forming a symmetric 3x3 matrix similar to~\cite{Atkeson1986}. The center of mass~$r_c$ is defined with respect to a body frame~$b$ and the inertia matrix~$H_{cm}$ is defined at the center of mass with the same orientation as the body frame.

This problem is novel from current methods for inertial estimation as,
\begin{enumerate}
\item They assume access to a force torque sensor, while we explore estimation with tactile fingertips that can only \HL{estimate} force.
\item They assume the object/load is  rigidly attached, while we do not attach the object to the fingertips and the object can move due to compliance existent in most common dexterous hands.
\item We are interested in estimating dynamics of low mass objects where the noise in sensing can have a significant impact on the estimation.
\item We also explore estimation of friction as friction plays a significant role in object dynamics when contacts can change.
\end{enumerate}

To explore this problem, our approach makes the following assumptions,
\begin{enumerate}
\item The object geometry is known as a mesh.
\item The contact between the robot's fingertips and the object is a point contact with friction~\cite{bicchi2000robotic}.
\item The object is a rigid object with a static mass distribution~(i.e., no moving parts).
\end{enumerate}

The proposed approach will work with any fingertips that can \HL{estimate} forces \HL{and contact point of the object on the fingertip}. In this paper, we focus on leveraging the BioTac tactile sensor to \HL{estimate} force \HL{and contact point} at fingertips. \HL{Though the BioTacs are soft fingertips, we select an in-hand grasp that avoids exertion of large frictional torques from the BioTacs on the object. We leave incorporating soft tactile fingertips in our formulation for the future as no method exists to estimate torque from the BioTac sensor.} Recently, a force estimation model for the BioTac was learned by~Sundaralingam~\etal~\cite{Sundaralingam-ICRA-19} using neural networks. We augment the learning architecture from~Sundaralingam~\etal~\cite{Sundaralingam-ICRA-19}  with multi-sample dropout layers~\cite{inoue2019multi} for the final fully connected layers. This augmentation enables us to get an uncertainty estimate of the force prediction without affecting inference speed. We encode this uncertainty estimate in our factor graph inference based approach to inertial estimation as we discuss later in Sec.~\ref{sec:inertial_est}. Leveraging this tactile force fingertips, we develop an interactive control scheme in Sec.~\ref{sec:fric_est} for estimating the friction. 

\subsection{Friction Estimation}
\label{sec:fric_est}
\begin{figure}[t]
  \centering
  \begin{tabular}{c c}    
    \includegraphics[trim={1cm 0.5cm 12cm 1cm},clip,height=4.5cm]{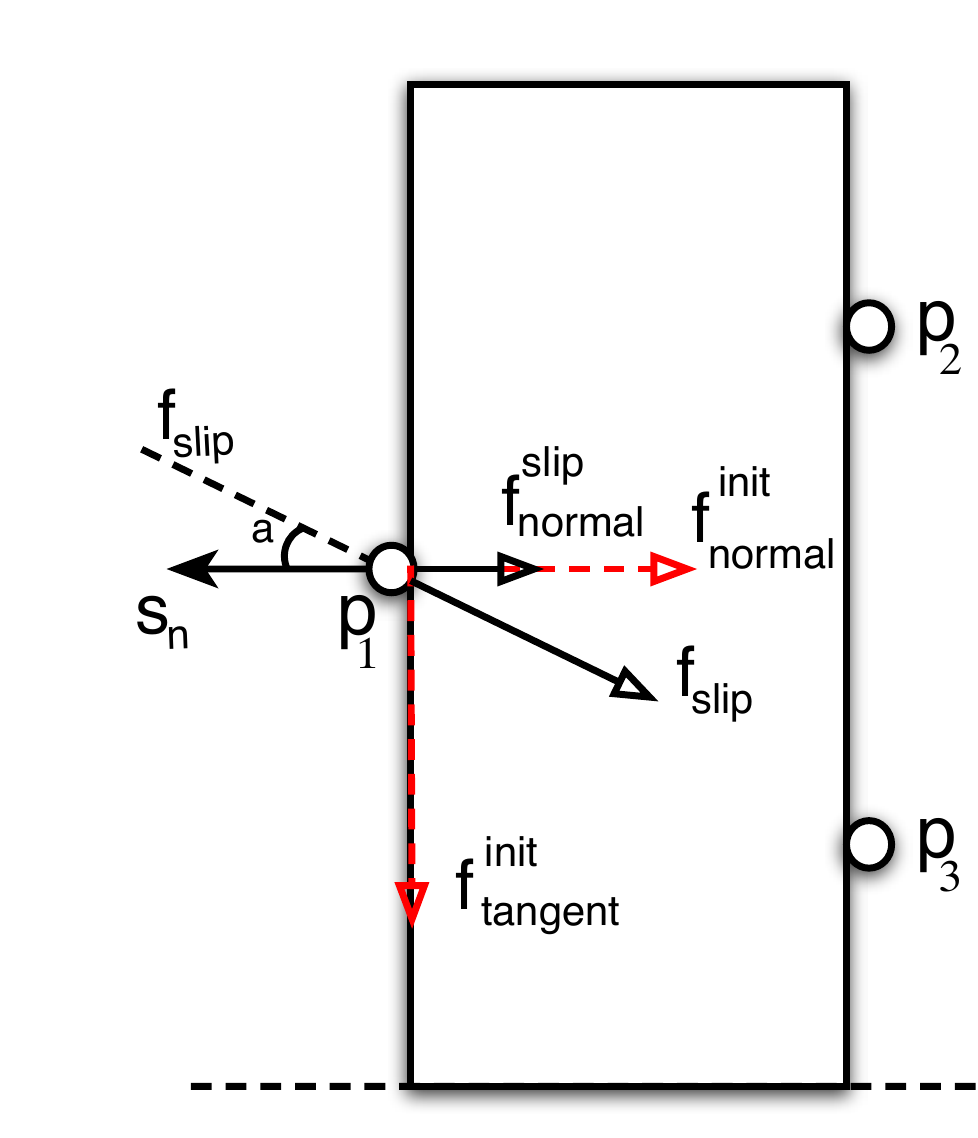}
    &     \includegraphics[trim={12cm 0.5cm 2cm 1cm},clip,height=4.5cm]{figs/notation}
\\
    (a) & (b)
    \end{tabular}
    \caption{The problem is illustrated in~$\mathbf{SE}(2)$ for clarity. The left figure~(a) and right figure~(b) are annotated for friction and inertial estimation respectively. The red vectors($f^{\textit{init}}_{\textit{normal}},f^{\textit{init}}_{\textit{tangent}}$ in~(a) represents the normal and tangent components of force applied initially on the surface. The force along normal direction is reduced until slip is detected and the normal component of the force before slip is shown as~$f^{\textit{slip}}_{\textit{normal}}$. In~(b), the center of mass is shown by the point~$C$ which is at a distance~$r_c$ from the object's body frame~$b$.}
  \label{fig:notations}
\end{figure}

To perform friction estimation, we leverage Coulomb's law. \HL{Though Coulomb's friction is an oversimplified model of real world friction, this simple friction model has been leveraged to perform many real world manipulation tasks~\mbox{\cite{shi2019inhand,chavan2018hand}}. According to this model,} a tangential force~$f_{\textit{tangent}}$ applied on an object surface is resisted by a frictional force~$f_{\textit{friction}}$ \HL{that acts tangential to the surface}. When the tangential force's magnitude is greater than the frictional force~$|f_{\textit{tangent}}|>\mu|f_{normal}|$, slip occurs. \HL{When a force~$f$ is applied at a contact~$p$ at an angle~$\alpha$ with respect to the surface normal~$s_n$ at the contact, this force can be split into two components, the normal force~$f_{\textit{normal}}=f\sin(\alpha)$ acting about the surface normal and the tangent force~$f_{\textit{tangent}}=f\cos(\alpha)$ acting tangential to the surface.} By reducing the normal force~$f_{\textit{normal}}$ and keeping the tangent force constant, we see the contact slowly reaching the slip boundary as shown in Fig.~\ref{fig:notations}. And the contact will slip once~$|f_{\textit{tangent}}|>\mu|f_{normal}|$. \HL{The force~$f_{slip}$ right before occurrence of slip is right at the maximum dry friction limit, giving us~$f_{tangent}=\mu f_{normal}$. We can derive an equation to estimate friction when the applied force is at this limit,}
\begin{align*}
  f_{tangent} &= \mu f_{normal}\numberthis \\
  f_{slip} \sin(\alpha) &= \mu f_{slip} \cos(\alpha) \numberthis\\
  \mu &= tan(\alpha) \numberthis \label{eq:mu}
\end{align*}
\HL{We can compute~$\alpha$ using the dot product between the unit force vector right before slip~$\hat{f}_{slip}$ and the surface normal~$s_n$ as,}
\begin{align*}
  \alpha&=\cos^{-1}(\hat{f}_{slip}\cdot s_n) \numberthis \label{eq:mu1}
\end{align*}
\HL{Substituting~Eq.~\mbox{\ref{eq:mu1}} in Eq.~\mbox{\ref{eq:mu}},}
\begin{align*}
  \mu&=\tan(\cos^{-1}(\hat{f}_{slip}\cdot s_n)) \numberthis \label{eq:friction_est}
\end{align*}
To perform friction estimation with a dexterous hand, the robot makes contact with the object using its~$n$ tactile fingertips and holds the object rigid. The normal component of the force applied by one of the tactile fingertips~$i$ is reduced slowly while the tangent component is kept constant until the fingertip slips. The net force just before slip~$f_{slip}$ is used to estimate the friction coefficient~$\mu$ using equation~\ref{eq:friction_est}.

\subsection{Inertial Estimation as Inference in a Factor Graph}
\label{sec:inertial_est}
To estimate the object's dynamic parameters~$[m,r_c,H_{cm}]$ from observed forces~\HL{$f_{i\in [1,n]}$} at contacts~\HL{$p_{i\in [1,n]}$} and the object's pose~$X_t$, we represent the problem as a factor graph and perform inference. The structure of our factor graph is shown in~Fig.~\ref{fig:factor_graph}. Measurements at timestep~$t$ of the object pose, contact points, and the contact forces from available perception is encoded into the factor graph with measurement factors to be~$X_t$, \HL{$p_{i,t}$} and~\HL{$f_{i,t}$} respectively. We will introduce our dynamics factor which relates the change in pose across timesteps, forces, contacts to the inertial parameters followed by the remaining factors for performing inertial inference.

\subsubsection{Dynamics Factor}

If forces~\HL{$f_{i\in[1,n]}$} are acting on the rigid object at positions~\HL{$p_{i\in[1,n]}$} with reference to the body frame, the newton-euler equations can be written as,
\begin{align*}
  \sum_if_i & =   m  (a - g) - m [r_c]_\times \dot{\omega} + m [\omega]_\times [\omega]_\times r_c  \numberthis \label{eq:f_eq}\\
    \sum_i [p_i]_\times f_i  &= m [r_c]_\times (a-g) +  H \dot{\omega}
            + [\omega]_\times H\omega  \numberthis \label{eq:tau_eq}
\end{align*}

where~$g$ is the gravity vector. The linear acceleration, angular acceleration, and angular velocity are represented as~$a$,~$\dot{\omega}$, and~$\omega$ respectively.

Since we do not assume to have object velocities and acceleration available from sensing, we compute them from observing the object pose~$\W{X}_{t-2}$, $\W{X}_{t-1}$, $\W{X}_{t}$ with reference to the world frame and the time~$T_{t-2}$, $T_{t-1}$, $T_{t}$ at timesteps~$t-2$, $t-1$, $t$ respectively. Additionally, we would like to estimate the inertial parameters in the body frame as the parameters are constant with respect to the body frame for any object motion. Rewriting Eq~\ref{eq:f_eq} and Eq.~\ref{eq:tau_eq} with reference to the body frame at timestep~$t-2$,
\begin{align*}
  \sum_i\T{f}_i & =   m  (\T{a} - \T{g}) - m [r_c]_\times \T{\dot{\omega}} \\ & + m [\T{\omega}_{t,t-1}]_\times [\T{\omega}_{t-1,t-2}]_\times r_c  \numberthis \label{eq:f_eq_b}\\
    \sum_i [\T{p}_i]_\times \prescript{t-2}{}{f}_i  &= m [r_c]_\times (\T{a}-\T{g}) +  \B{H} \T{\dot{\omega}}
            \\ & + [\T{\omega}_{t,t-1}]_\times\B{H}\T{\omega}_{t-1,t-2}  \numberthis \label{eq:tau_eq_b}
\end{align*}
where the superscript~$t-2$ refers to body frame at timestep~$t-2$.

We will now derive the linear acceleration~$a$, angular acceleration~$\dot{\omega}$ and the angular velocity~$\omega$ from the pose of the object through finite difference. Given the poses~$\W{X}_{t-2}$, $\W{X}_{t-1}$, $\W{X}_{t},$ of the object in the world frame at time~$t-2$, $t-1$, $t$, the relative pose change between timesteps can be written as,
\begin{align*}
  \T{X}_{t-1} & =   \prescript{w}{}{X}_{t-2}^{-1}  \prescript{w}{}{X}_{t-1}\numberthis\\
  \prescript{t-1}{}{X}_{t} & =   \prescript{w}{}{X}_{t-1}^{-1}  \prescript{w}{}{X}_{t} \numberthis
\end{align*}
\HL{If the sampling time is very small~(i.e.,~$T-T_{t-1}$ is very small), we can assume a constant velocity from~$t-1$ to~$t$. Then,} the instantaneous velocities can be computed following~\cite{Forster-tro} as,
\begin{align*}
  \T{\dot{X}}_{t-1,t-2} &=   \frac{\log(\T{X}_{t-1})}{T_{t-1}-T_{t-2}} \numberthis \\
  \prescript{t-1}{}{\dot{X}}_{t,t-1} &=   \frac{\log(\prescript{t-1}{}{X}_{t})}{T_{t}-T_{t-1}} \numberthis
\end{align*}
where~$T_{t-2}$, $T_{t-1}$, $T_{t}$ are clock time at timesteps~$t-2$, $t-1$, $t$ respectively. The angular velocities~$\omega_{t-1,t-2}$, $\omega_{t,t-1}$ in Eq.~\ref{eq:f_eq_b} and Eq.~\ref{eq:tau_eq_b} are the angular components of~$\T{\dot{X}}_{t-1,t-2},  \prescript{t-1}{}{\dot{X}}_{t,t-1}$ respectively.

To compute the acceleration, we first transform the velocity~$  \prescript{t-1}{}{\dot{X}}_{t,t-1}$ to the body frame at timestep~$t-2$,
\begin{align*}
    \T{\omega}_{t-1,t} & =   \prescript{t-1}{}{\omega}_{t-1,t} +    \T{\omega}_{t-2,t-1} \times \prescript{t-1}{}{\omega}_{t-1,t} (T_{t}-T_{t-1}) \numberthis \\                                
  \T{v}_{t-1,t} & =   \prescript{t-1}{}{v}_{t-1,t} +    \T{\omega}_{t-2,t-1} \times \prescript{t-1}{}{v}_{t-1,t} (T_{t}-T_{t-1}) \numberthis
\end{align*}
where $\omega$ and $v$  are the angular and linear components of~$\dot{X}$ respectively. The linear acceleration~$a$ and the angular acceleration~$\dot{\omega}$ can be computed as
\begin{align*}
  {\dot{\omega}} &= \frac{\T{\omega}_{t-1,t} - \T{\omega}_{t-2,t-1} }{T_{t}-T_{t-1}} \numberthis \\
    {a} &= \frac{\T{v}_{t-1,t} - \T{v}_{t-2,t-1} }{T_{t}-T_{t-1}} \numberthis
\end{align*}

The dynamics factor~$D(\cdot)$ can be written as,
\begin{align*}
  D(f_{i,t-2},p_{i,{t-2}},H,&\W{X}_t,\W{X}_{t-1},\W{X}_{t-2}) \\ &=\begin{bmatrix}
A -       \sum_i\T{f}_i \\
   B -\sum_i [\T{p}_i]_\times \prescript{t-2}{}{f}_i
 \end{bmatrix}
\end{align*}
where,
\begin{align*}
  A&=    m  (a - \T{g}) - m [r_c]_\times \dot{\omega} + m [\omega_{t,t-1}]_\times [\omega_{t-1,t-2}]_\times r_c \numberthis \\
  B&= m [r_c]_\times (a-\T{g}) +  \B{H} \dot{\omega}
   + [\omega_{t,t-1}]_\times\B{H}\omega_{t-1,t-2} \numberthis
\end{align*}

\subsubsection{Geometric Prior Factor}
\label{sec:geom-prior-fact}
The inertial parameters are in a positive definite manifold as shown by~\cite{lee2018geometric}. We leverage their work and encode a prior~$w_0$ on the inertial parameters with a factor~$B(W)$ that minimizes the geodesic distance on the Reimmannian manifold between the current estimate of~$W$ and the prior~$w_0$.
\begin{align*}
  B(W) &= \Tr(p_{0}^{-1} p) \numberthis \label{eq:geodesic}
\end{align*}
where~$p,p_{0}\in P(4)$ are projection of~$w,w_0$ into the positive definite manifold. This is done using the following function,
\begin{align*}
  \text{proj}(m,r_c,H)&= \begin{bmatrix}
    \frac{1}{2} \Tr(H) I_{33}-H & m r_c\\
    m r_c^\top & m
  \end{bmatrix}\numberthis \label{eq:geo_project}
\end{align*}

\begin{figure}
  \centering \includegraphics[width=0.48\textwidth]{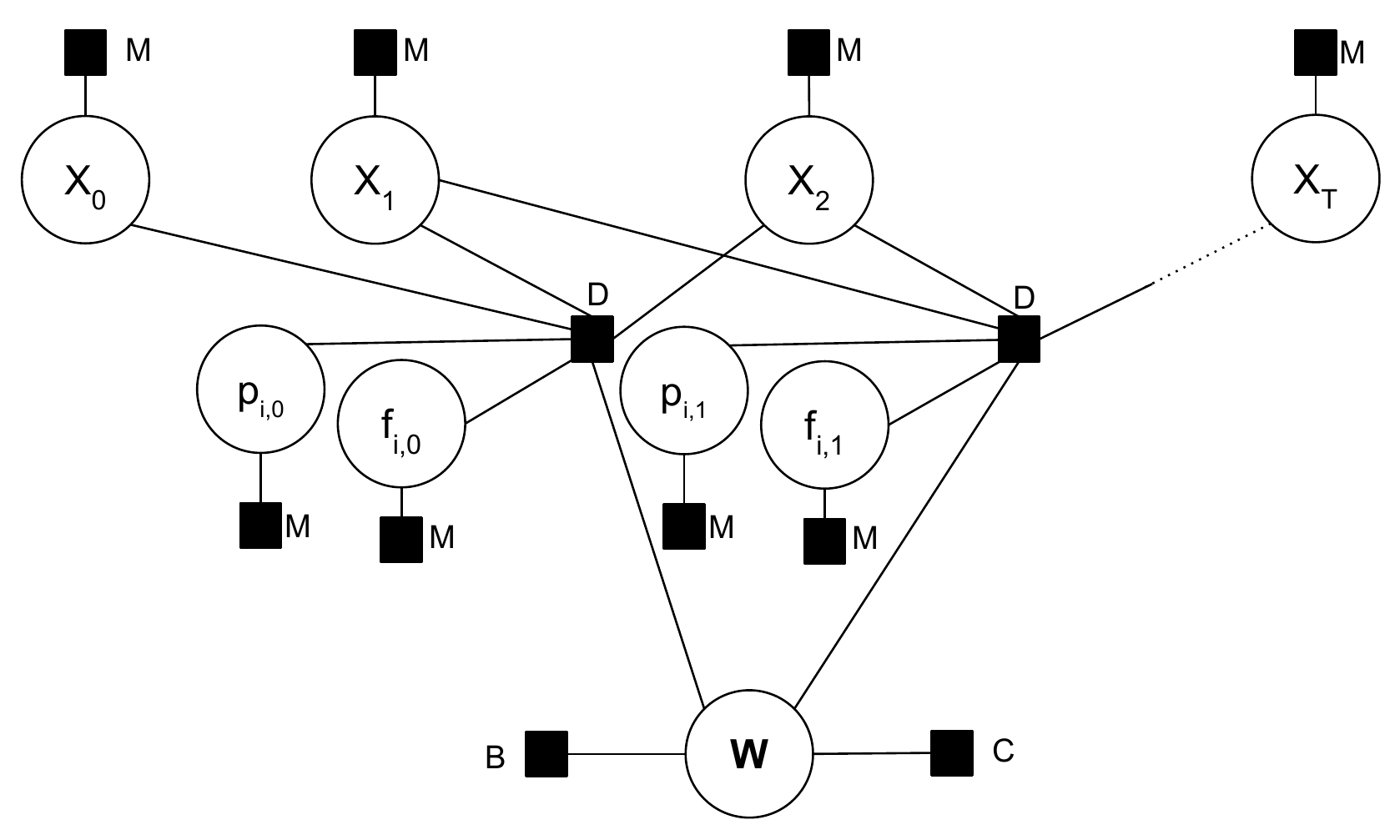}
  \caption{Illustration of variables and the associated factors of our proposed approach for object dynamics inference. Circles are variables and squares are factors. Factors~M, D, C,~and B represent measurement, dynamics, constraint, and inertial prior factors respectively.}
\label{fig:factor_graph}
\end{figure}

\subsubsection{Physical Consistency Factor}
\label{sec:inequality}
We leverage the work done by Traversaro~\etal~\cite{traversaro2016identification} to enable physical consistency of the inertial matrix. The inertial matrix of a rigid body about its principle axes is a positive diagonal matrix,
\begin{align*}
  H_{diag}&= \int_{r \in O} [r]_\times [r]_\times \rho(r) dr \numberthis
\end{align*}
where~$\rho(r)$ is the density of the particle~$r$ in the object~$O$ and each component of the diagonal matrix~$H_{diag} = \text{diag}(H_x,H_y,H_z)$ being,
\begin{align*}
  H_x &= \int \int_{\mathbb{R}^3} \int (r_y^2 + r_z^2) \rho(r) dr \numberthis \\ 
  H_y &= \int \int_{\mathbb{R}^3} \int (r_x^2 + r_z^2) \rho(r) dr \numberthis \\
  H_z &= \int \int_{\mathbb{R}^3} \int (r_x^2 + r_y^2) \rho(r) dr \numberthis
\end{align*}
Defining~$L_x = \int \int_{\mathbb{R}^3} \int (r_x^2) \rho(r)dr $, $L_y = \int \int_{\mathbb{R}^3} \int (r_y^2) \rho(r)dr$, and $L_z = \int \int_{\mathbb{R}^3} \int (r_z^2) \rho(r)dr$, enables writing~$H_x,H_y,$ and~$H_z$ as,
\begin{align*}
  H_x &= L_y + L_z \numberthis \\
  H_y &= L_x + L_z \numberthis \\
  H_z &= L_x + L_y \numberthis
\end{align*}
and makes the inertial matrix~$H_{diag}$ about the principal axes to be,
\begin{align*}
  H_{diag} &= \text{diag}(L_y+L_z, L_x+L_z,L_x+L_y) \numberthis \label{eq:h_const}
 \end{align*}
 To guarantee positive definiteness of~$H_{diag}$, it is sufficient if ~$L_{x,y,z} \succ 0$ holds. Now, we can transform this diagonal inertial matrix~$H_{diag}$ to be about any frame with origin at the center of mass and rigidly fixed to the object by using a rotation matrix,
\begin{align*}
  \B{H} & = \B{R}_a H_{diag}  \B{R}_a^\top \numberthis
\end{align*}
where~$\prescript{b}{}{R}_a$ is the rotation matrix from the principle axes frame~$a$ to the body frame~$b$. By parallel axis theorem, we can recover the inertial matrix~$H_{cm}$ at the center of mass as~$H_{cm}= \B{H} - m r_c r_c^\top$. The inertial matrix~$\B{H}$ can now be represented with three real numbers~$[L_x,L_y,L_z]\in \mathbb{R}$ and a rotation matrix~$\B{R}\in \mathbf{SO}(3)$.

We will now derive bounds on the mass and center of mass parameters. If mass of a particle~$x$ that is in the object is given by~$M(x)$, the total mass~$m$ of an object can be written as,
\begin{align*}
  m&=\int_{x \in O} M(x) dx \numberthis \label{eq:mass_eq}
\end{align*}
For physical objects in the real world, the mass is positive everywhere on the object,
\begin{align*}
  M(x) > 0, \forall x \in O \numberthis \label{eq:mass_x_const}
\end{align*}
this makes~$m$ also to be always positive~($m>0$).

If~$p(x)$ defines the location of the particle with reference to the object's origin frame, the center of mass for an object can be defined as,
\begin{align*}
  r_c &= \frac{1}{m} \int_{x\in O} M(x) p(x) dx \numberthis \label{eq:com_1}
\end{align*}
making normalized mass function~$M'(x) = \frac{M(x)}{m}$,
Substituting~Eq.~\ref{eq:mass_eq} in Eq.~\ref{eq:com_1},
\begin{align*}
  r_c &= \int_{x \in O} M'(x) p(x) dx \numberthis \label{eq:com}
\end{align*}
Now~$\int_{x \in O} M'(x) dx = 1$ as~$0 < M'(x) \leq 1, \forall x \in O$. Eq.~\ref{eq:com} resembles a convex combination function. If the set~$O$ is convex then~$r_c$ also lies inside this convex set. For non-convex objects, the center of mass~$r_c$ will be inside the convex hull of the object~$CVX(O)$.

The inertial parameters~$m,r_c,H$ are rewritten using the above discussed variables as~$w = [m,r_{c},L_x,L_y,L_z,\B{R}_a ]$, where~$[m,L_x,L_y,L_z]\in \mathbb{R}, r_c\in \mathbb{R}^3$ and~$\B{R}_a\in \mathbf{SO}(3)$. Consolidating all our physical consistency constraints,
\begin{align*}
  m & > 0\numberthis \label{eq:m_ineq} \\
  r_c &\in CVX(O) \numberthis \label{eq:c_ineq}\\          
  L_x,L_y,L_z &\succeq 0 \numberthis \label{eq:h_ineq}
\end{align*}
These constraints are encoded by the factor~$C(w)$ as shown in fig.~\ref{fig:factor_graph}.

%%% Local Variables:
%%% mode: latex
%%% TeX-master: "../paperdraft"
%%% End:

\section{Evaluation Details}
\label{sec:implement}
To validate our proposed approach, we perform extensive evaluation on a real robot platform and also in simulation. We first describe our setup and data collection process in Sec.~\ref{sec:data-collection}, followed by our approach to comparing with other methods in Sec.~\ref{sec:comparison}.
\begin{figure}
  \centering
  \begin{tabular}{c c}
    \includegraphics[trim={0 1cm 0 1cm},clip, width=0.45\columnwidth]{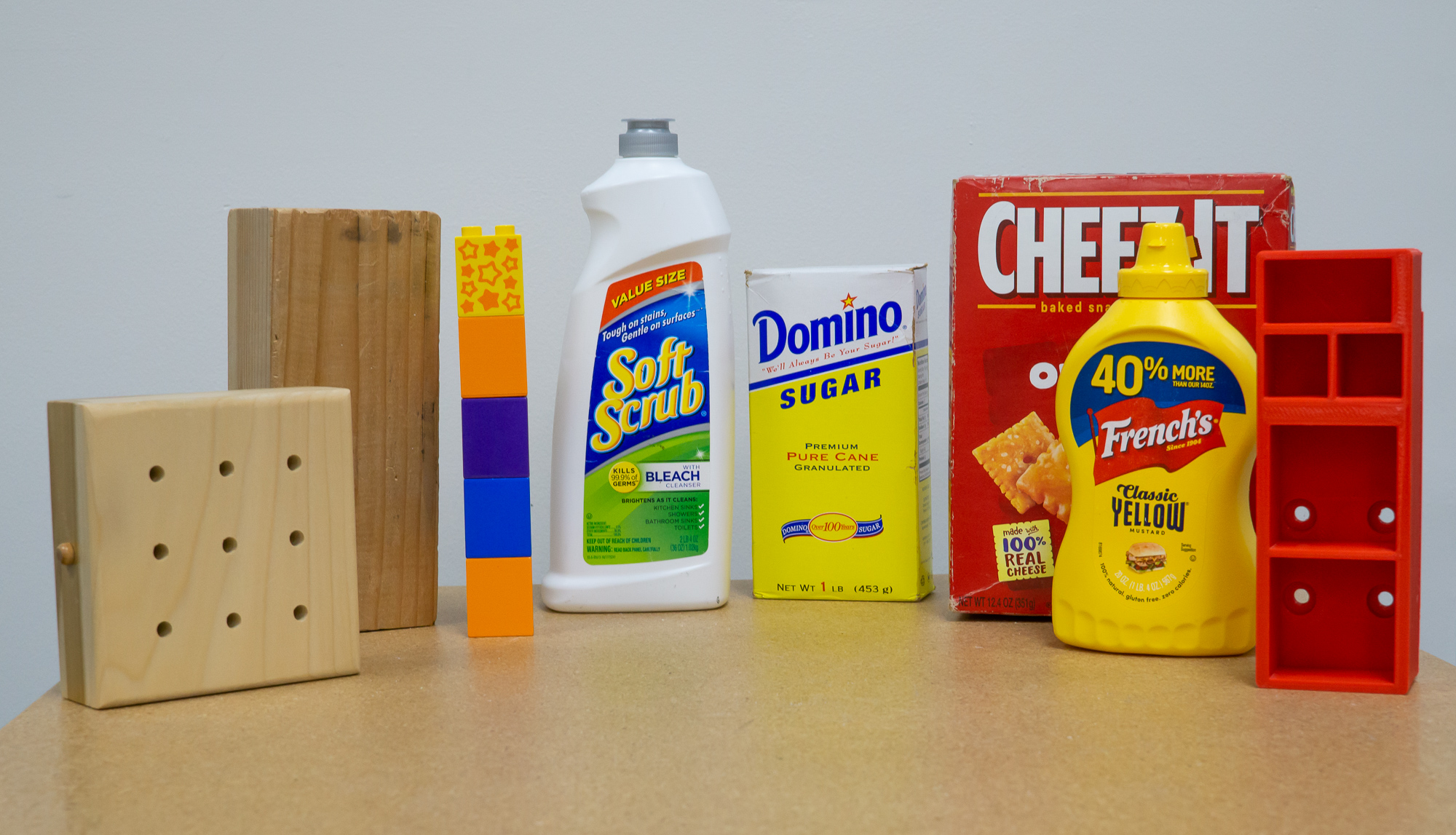} &                                                                                                    \multirow{3}{*}[1.6cm]{\includegraphics[trim={1cm 1cm 0cm 1cm},clip,width=4cm]{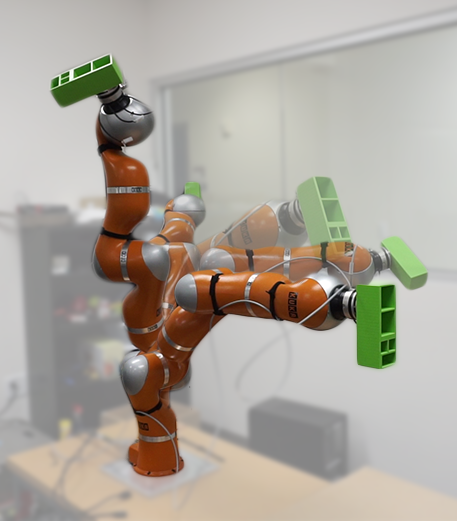}}
    \\
    (a) Friction Objects & \\
    \includegraphics[trim={0 5cm 0cm 4cm},clip,width=0.45\columnwidth]{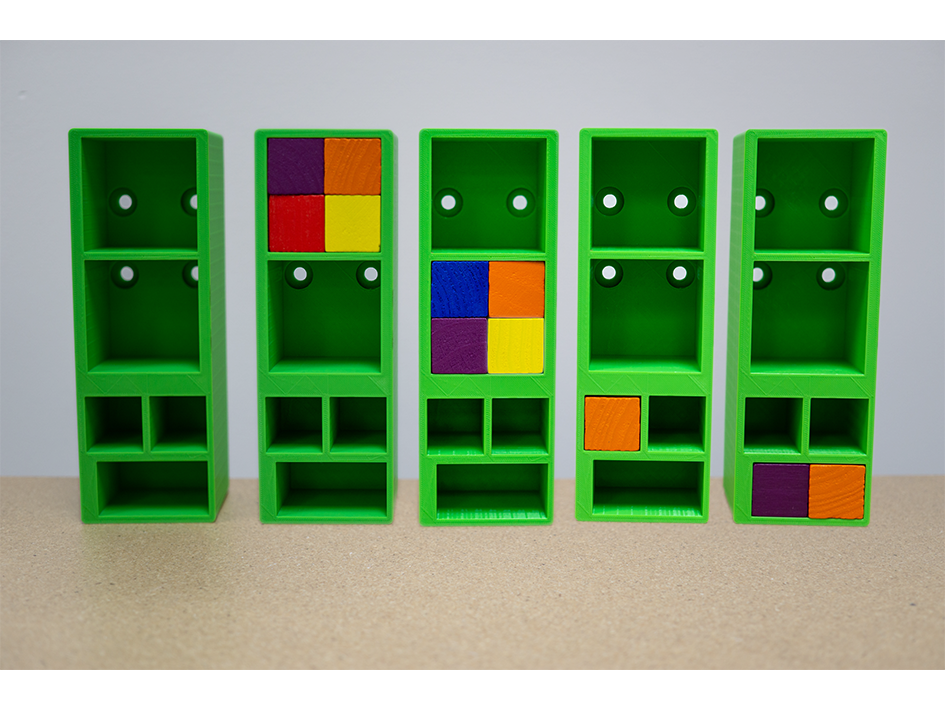} & \\
    (b) Inertial Objects & (c) \HL{Wrist-FT} estimation
  \end{tabular}
  \caption{Objects from the YCB dataset with labels 9 hole peg, wood block, Lego Duplo, bleach cleanser, box of sugar, cracker box, mustard container~(from left to right) and a 3D printed object are used to validate friction estimation. A 3D printed object~(the left most) is augmented with blocks to validate inertial estimation. The objects are named~box\_0 to box\_4 from left to right.}
  \label{fig:objects}
\end{figure}

\subsection{Implementation, Experiments, \& Data Collection }
\label{sec:data-collection}
% Robot setup
Our real world experimental setup consists of an Allegro hand with BioTac tactile fingertips, mounted on a KUKA LBR4 arm. An OptoForce HEX-E 6-DOF force torque sensor is mounted at the wrist between the arm and the hand. Visual perception is enabled by an ASUS Xtion camera and pose tracking of the robot and the object is done using Dense Articulated Realtime Tracker~(DART)~\cite{schmidt2015depth}.

% friction estimation
To command force for friction estimation, we setup a Cartesian impedance controller using the transpose of the kinematic Jacobian to map the force to a joint position command. We use the index finger to estimate friction while the other fingertips hold the object rigidly as shown in Fig.~1. We find the nearest triangle to the contact point on the mesh using~KrisLibrary~\cite{hauser2016robust} and use this triangle's surface normal as~$s_n$. To validate friction estimation, we select 8 objects~as shown in Fig.~\ref{fig:objects}-(a) from the YCB dataset~\cite{calli2015}. For~\emph{mustard} and \emph{bleach cleanser} objects, we perform friction estimation on both plastic and label regions. We conduct 10 trials per object.

\HL{We use the slope method~\mbox{\cite{slope_friction_url}} to compute the ground truth coefficient of friction. We mount the object on a platform which can rotate the object with respect to the gravity vector. Then, we place the BioTac on the object and track them using vision. We then rotate the platform until the BioTac slips and record the angle between the gravity vector and the surface normal when this slip happens. Using equation~\mbox{\ref{eq:friction_est}} and substituting the unit gravity vector with reference to the object frame as ~$\hat{f}_{slip}$ and the surface normal at the contact point of the BioTac as~$s_n$, we get the ground truth friction coefficient. We repeated this experiment 10 times and take the mean value as the ground truth friction coefficient for each object.}

We implement the factor graph in GTSAM~\cite{dellaert2012factor} and run batch optimization using Levenberg Marquardt optimizer. To use inequality constraints from equations~\ref{eq:m_ineq}-\ref{eq:h_ineq} in GTSAM, we set them to be active only when the constraint is violated. To validate inertial estimation, we test on three data sources:
\begin{description}[leftmargin=0cm]
\item[Simulation:] We simulate an object with four points of force application in dartsim~\cite{Lee2018}. This helps us validate our approach on an highly inertial object without worrying about joint torque limits. This also helps us tune the weights of the different factors. We add noise from a zero mean normal distribution with variances~$\sigma^2=[0.1,0.25,0.5,1.0]$ to the forces, to study the robustness of our method.
\item[Wrist-FT:] We rigidly attach the object to a wrist mounted force torque  sensor and move along different trajectories as shown in Fig.~\ref{fig:objects}-(c). This setup is similar to Atkeson~\etal~\cite{Atkeson1986} and shows how our proposed approach can be used to perform load inertial estimation. We leverage the joint angle sensors of the arm to compute the object pose as it is rigidly attached.
\item[In-Hand:] The robot grasps the object with its four tactile fingertips and moves along trajectories as shown in Fig.~1. This setup resembles how a robot could use our approach for inertial estimation in the unstructured, real world. When the object is grasped in-hand, we found depth based tracking to be very poor as the pointcloud of the object wasn't sufficient rich. Hence, we also explore tracking the object kinematically as a rigid attachment to the thumb following our previous work on ``Relaxed-Rigidity'' constraints for in-grasp manipulation~\cite{sundaralingam2019relaxed}. For kinematic tracking, we initialize the rigid frame between the thumb and the object from~DART and use joint angle sensing to track the object.
\end{description}

For Wrist-FT and In-Hand, we 3D print an object and augment it with wooden blocks as shown in Fig.~\ref{fig:objects}-(b). We compute the ground truth mass using a weigh scale. We compute ground truth values of inertia and center of mass by using CAD. We tune the weights of the factors from no noise simulation data and then keep them fixed across all our data sources to enable direct comparison across methods and data sources.

\subsection{Error Metrics \& Comparison Methods}
\label{sec:comparison}
% Error metric for friction
We compute the difference between the ground truth~$\mu_{gt}$ and the estimated friction~$\mu$ as ``Friction Estimation Error''~$=\mu_{gt}-\mu$ to analyze if our estimation under estimated or over estimated friction. We also report the estimated friction to study performance at different friction ranges.

% inertial estimation
We report the estimated inertial parameters in a numeric table as supplementary material. We also use the projected ground truth~$p_{gt}$ and estimated~$p_{est}$ inertial parameters to compute the``inertial error'' as,
\begin{align*}
  \text{inertial error}&= |4-\Tr(p_{gt}^{-1} p_{est})| \numberthis 
\end{align*}
where~$p_{gt}$ and~$p_{est}$ are projected into the inertial manifold using~Eq.~\ref{eq:geo_project}. This ``inertial error'' is the geodesic distance between the ground truth and estimated inertia as shown by~\cite{lee2018geometric}. This error metric is scale invariant and gives us a scalar value to interpret the error in inertial estimation across different methods.

We compare our inertial estimation method against the least squares approach introduced by~Atkeson~\etal~\cite{Atkeson1986}. We term this method as~``Baseline''.
In ``Baseline'', the optimization is directly over the inertial parameters~$w_{\textit{vector}}=[m,r_c,H_{xx},H_{yy},H_{zz},H_{xy},H_{xz},H_{yz}]$, assuming them to be independent variables in the vector space. We do not optimize over the sensor measurements of object pose, forces, and contacts. We smooth the force torque measurements with a Savitzky-Golay filter~(cubic polynomial with window size of 51) as without the filter the optimization failed.

We also implement a factor graph version of~\cite{Atkeson1986} and call this method~``Baseline-FG''. In ``Baseline-FG'', the optimization is over the vector of inertial parameters~$w_{\textit{vector}}$ and the sensor measurements of object pose, forces, and contacts. We also disable different factors in our approach to study their effect in estimation, specifically:
\begin{description}[leftmargin=0cm]
\item[No Constraint, No Geodesic] This version of the factor graph does not have the physical consistency factor~$C$ and the Geodesic prior factor~$B$. This version is a factor graph version of~\cite{traversaro2016identification} with sensor observations.
\item[Constraint, No Geodesic] This version does not have the Geodesic prior factor~$B$. This enables studying inertial inference with no initial estimates.
\item[Constraint+Geodesic] This version contains all the factors introduced in our approach. We set the prior to be the ground truth inertial parameters. This version is used to study the effect of incorporating initial estimates of the inertial parameters.
\end{description}

For all our factor graph based methods, we do not perform any pre-processing of the sensor measurements~(e.g., no filtering/smoothing of the sensor signals) as we found our structured approach estimated better directly from the raw signals.

%%% Local Variables:
%%% mode: latex
%%% TeX-master: "../paperdraft"
%%% End:

\section{Results}
\label{sec:results-discussion}
We report the results of our approach to object dynamics inference. In all box plots the middle line in the box plot defines the median error. The bottom and top borders indicate the first and third quartiles. The whiskers indicate the extrema of the inliers within 1.5 times the interquartile range.

\subsection{BioTac Force Estimation}
\label{sec:biot-force-estim}
\begin{figure}[t]
  \centering
  \includegraphics[width=0.49\textwidth]{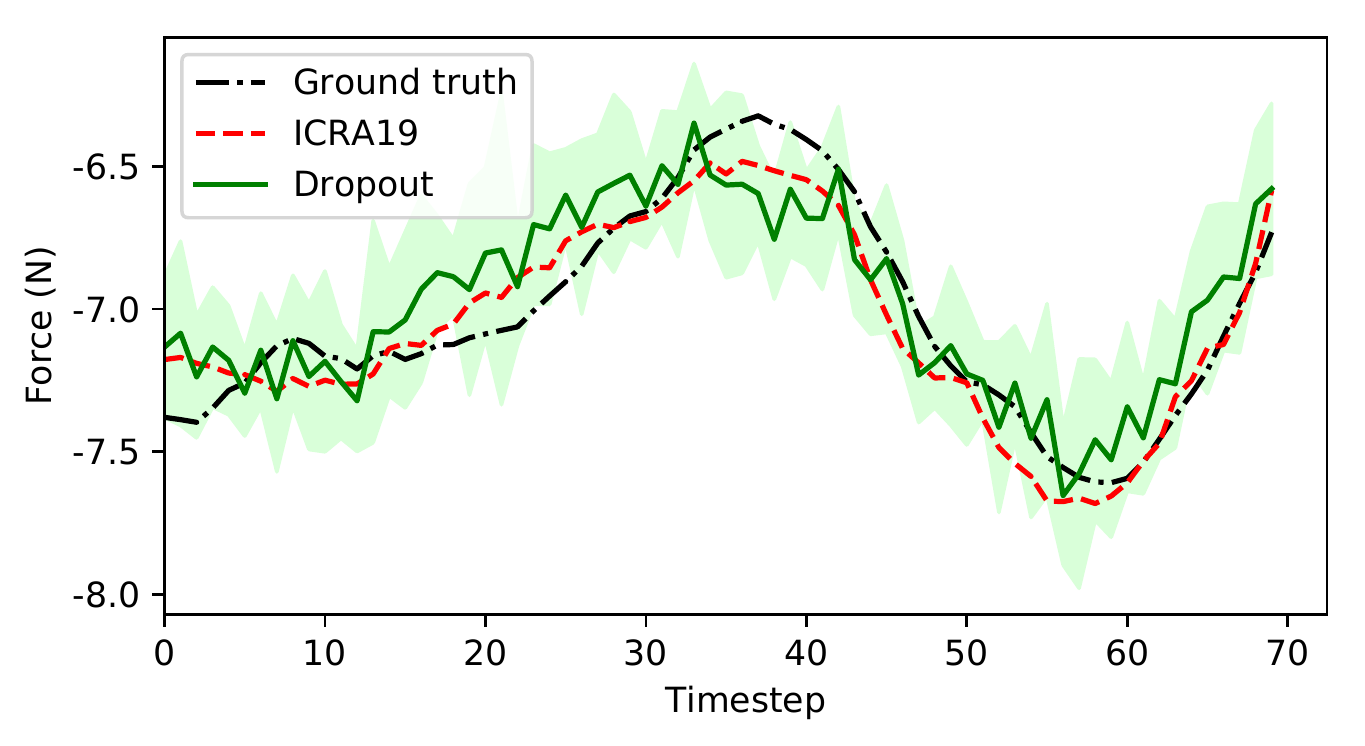}\\
  \includegraphics[width=0.49\textwidth]{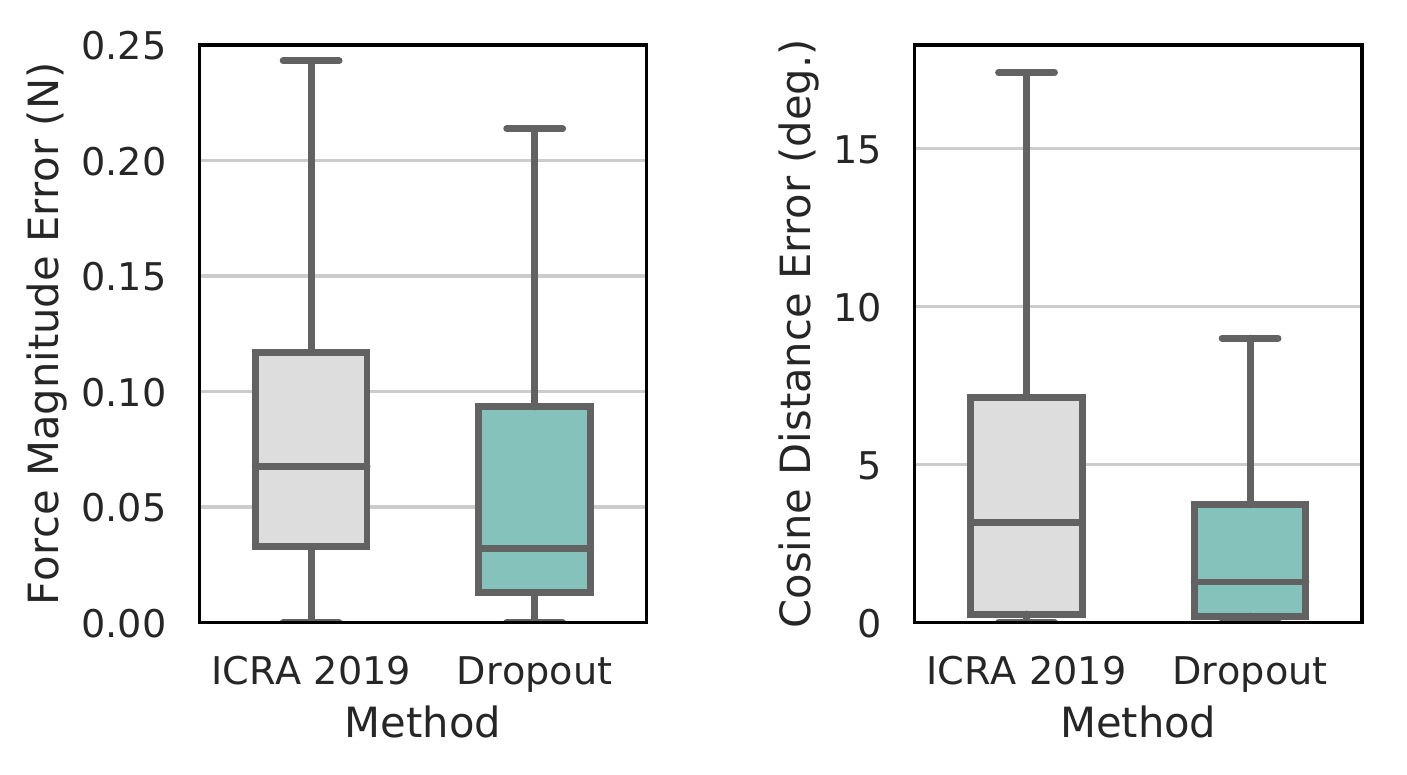}
  \caption{The force prediction using multi-sample dropout with one standard deviation of uncertainty is shown in the top plot for the $z$-axis \HL{along with the ground truth force and the force estimate from ICRA~2019~\mbox{\cite{Sundaralingam-ICRA-19}}}. The bottom plot shows the improved accuracy of force estimation with multi-sample dropout~(``Dropout'') compared to the existing method~ICRA~2019~\cite{Sundaralingam-ICRA-19}.}
\label{fig:force_plot}
\end{figure}
\HL{Across the test dataset from~\mbox{\cite{Sundaralingam-ICRA-19}} we found the ground truth force fell mostly within one standard deviation of the uncertainty estimate from the multi-sample dropout method as seen in~Fig.~\mbox{\ref{fig:force_plot}}}. The augmentation with multi-sample dropout also improved the performance of the learned model with a median magnitude accuracy of~$0.03N$ and an angular accuracy of~$1.28$ degrees which is 2x improvement over the current best method~``ICRA 2019''~\cite{Sundaralingam-ICRA-19}.

\subsection{Friction Estimation}
\begin{figure}[t]
  \centering
  \includegraphics[width=0.49\textwidth]{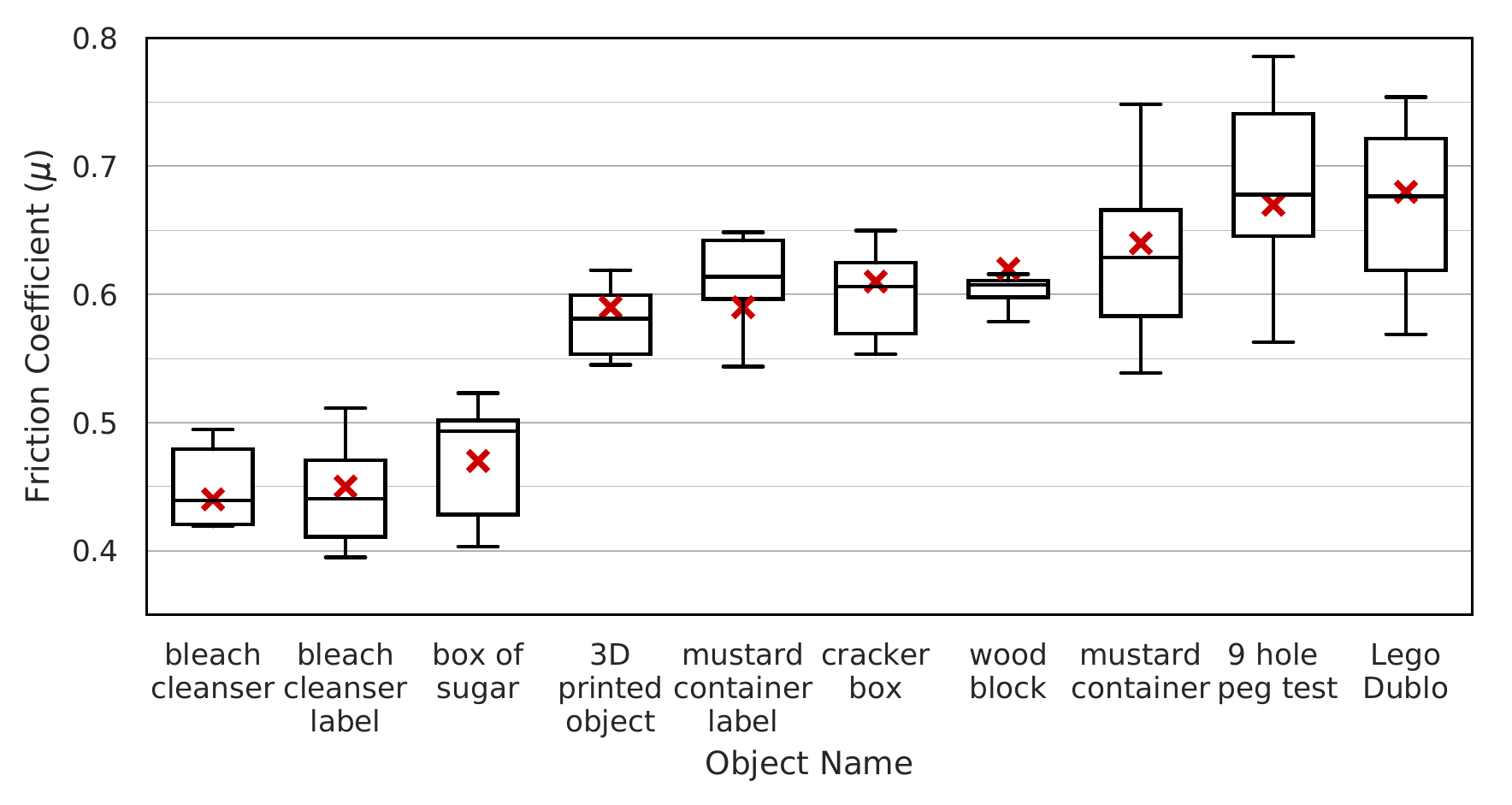}\\
  \includegraphics[width=0.49\textwidth]{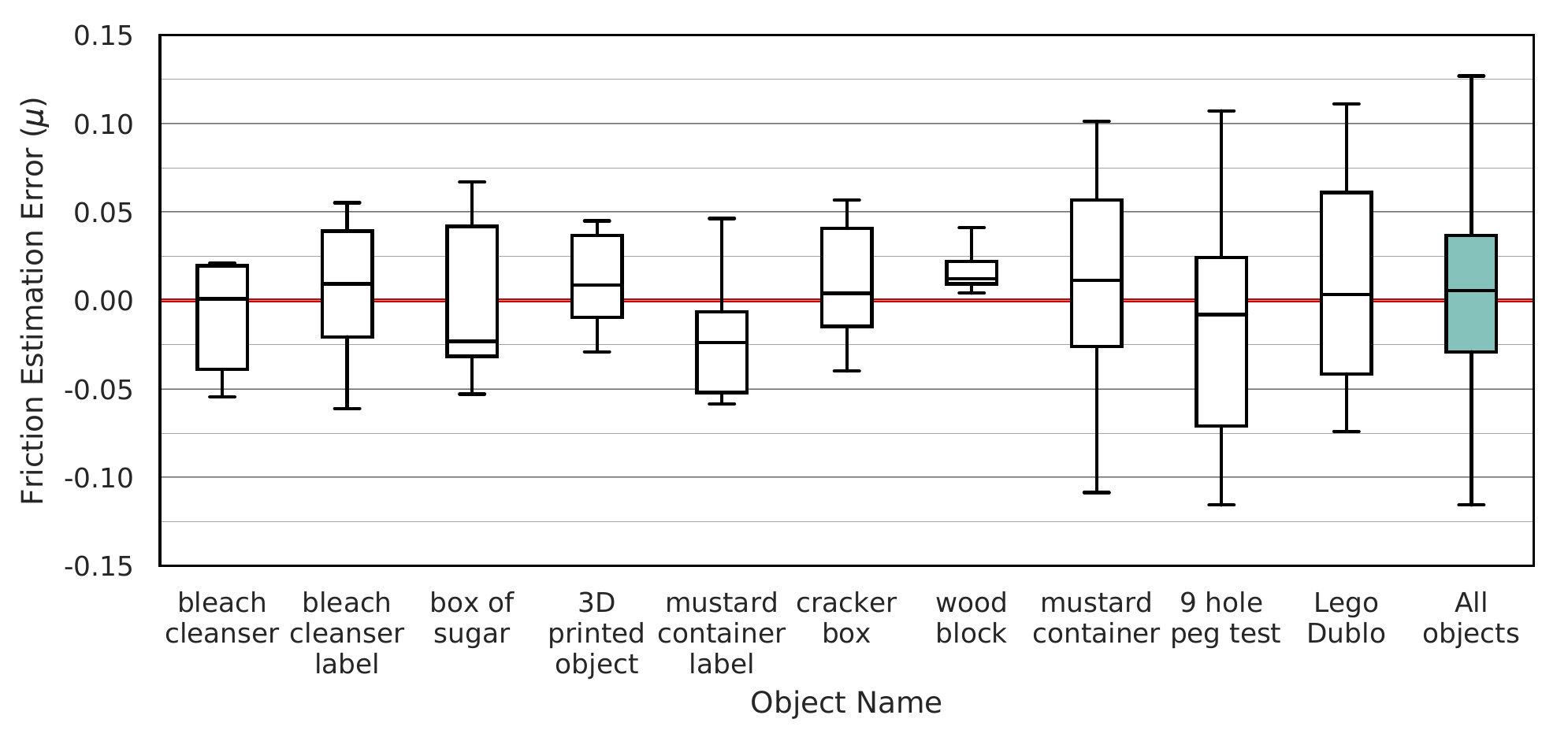}\\
  \caption{Estimated friction from our method is shown with the ground truth as the red ``x'' in the top plot. The bottom plot shows the difference between the ground truth and the estimated \HL{friction}. The suffix~\emph{label} to the object name to signifies the label region of the object.}
\label{fig:friction_plot}
\end{figure}

As we see in Fig.~\ref{fig:friction_plot}, the medians of our estimates for the friction coefficient are very close to the ground truth except for the \emph{box of sugar} and~\HL{\emph{mustard container label}} objects where we saw the object warping when force was applied by the fingertip. \HL{For the non-label surface of the \emph{mustard container} object, we didn't see any warping and this reflects in the improved accuracy of the estimated friction.} The median error between the estimated friction \HL{coefficient} and ground truth friction \HL{coefficient} across all objects was~$0.0056$. 

\subsection{Inertial Estimation}
\label{sec:inertial-estimation}
% TODO: cleanup results
\begin{figure}[t]
  \centering \includegraphics[trim={0 0cm 0 1.5cm},clip,width=0.49\textwidth]{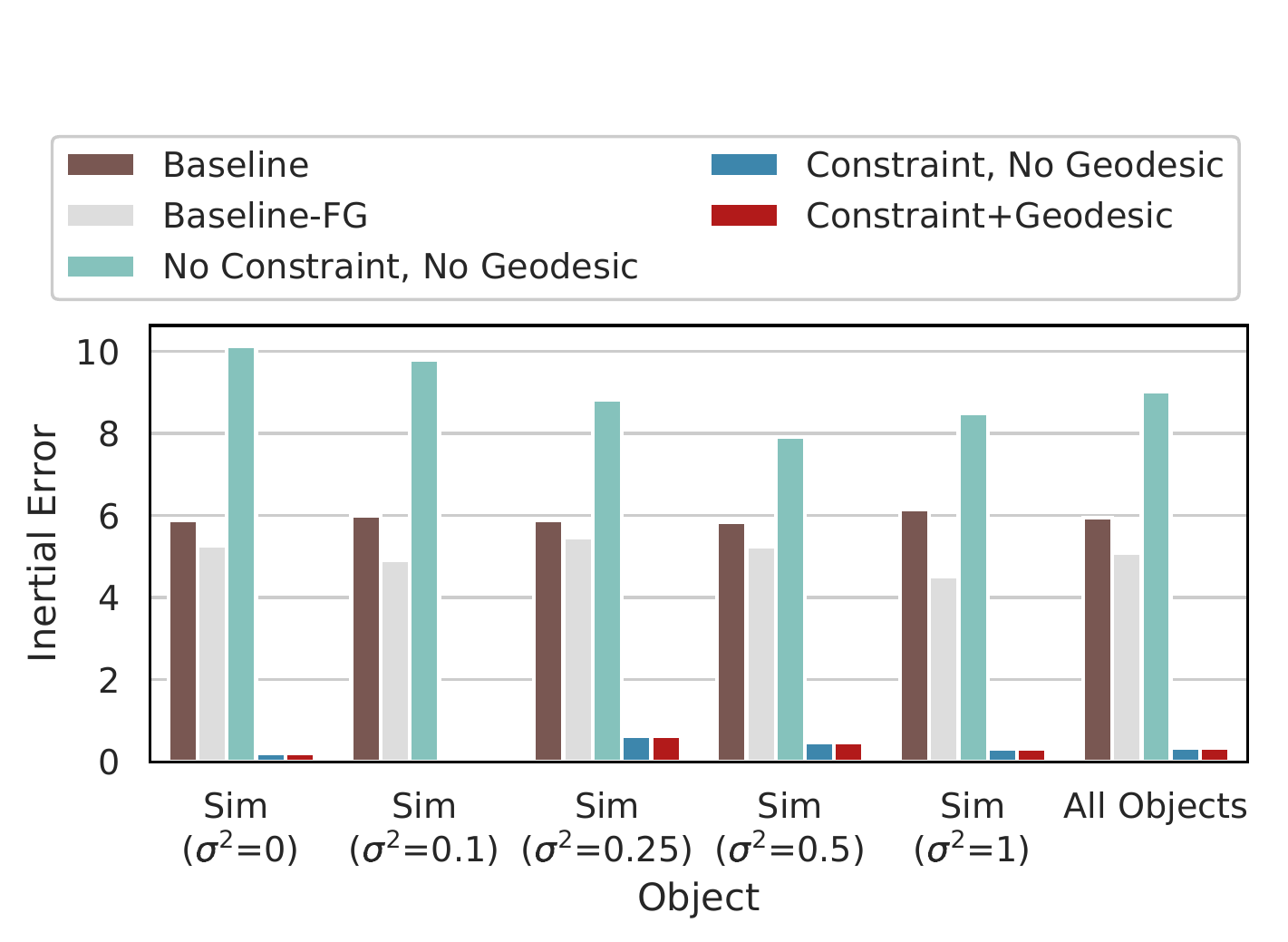}\\ \includegraphics[width=0.49\textwidth]{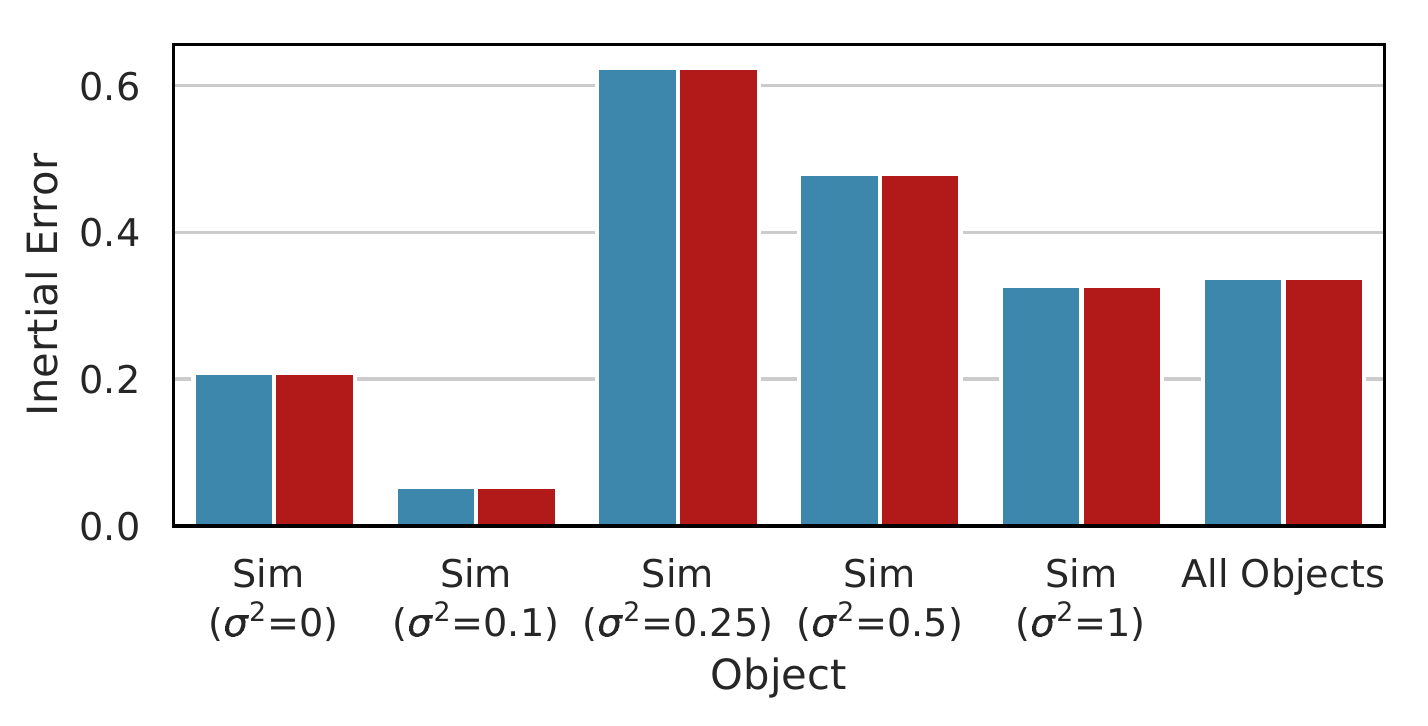}\\
  \caption{The inertial error in simulation is shown with different additive noise variances on forces. We show all methods in the top plot and show our two best methods in the bottom plot.}
\label{fig:sim_inertial}
\end{figure}

To accurately estimate all the inertial parameters, the object motion needs to sufficiently excite each of the inertial parameters. In this paper, we only focus on recovering the inertial parameters given fixed trajectories of the object and do not focus on active excitation of the inertial parameters for accurate estimation. Hence, the estimated inertial parameters will be numerically different from the ground truth. We could tune the weights to obtain numerically similar values for some of the inertial parameters. However, numerical similarity in some parameters will skew the other parameters to be further away from the object's true dynamics. Hence, we focus on estimation that is closer to the object's dynamics than being numerically similar. We do note that across all the data sources, our method~(``Constraint, No Geodesic'') found numerically similar values for the diagonal elements of the inertial matrix.  

% talk about computation time
In simulation, we recover the inertial parameters sufficiently well as shown in Fig.~\ref{fig:sim_inertial}. The recovered inertial parameters are also similar numerically as the object motion sufficiently excites most of the inertial parameters. We observed that even with zero noise the ``Baseline'' approach estimates the mass very inaccurately as the assumption of constant velocity between timesteps fails to hold. When evaluating ``Baseline'' on the other data sources, we found the optimization to fail or found very bad estimates such as negative mass. We hence do not report ``Baseline'' on the other data sources.

\subsubsection{Wrist-FT}
\label{sec:inert-estim-wrist}
\begin{figure}[t]
  \centering \includegraphics[trim={0 0cm 0 1cm},clip,width=0.49\textwidth]{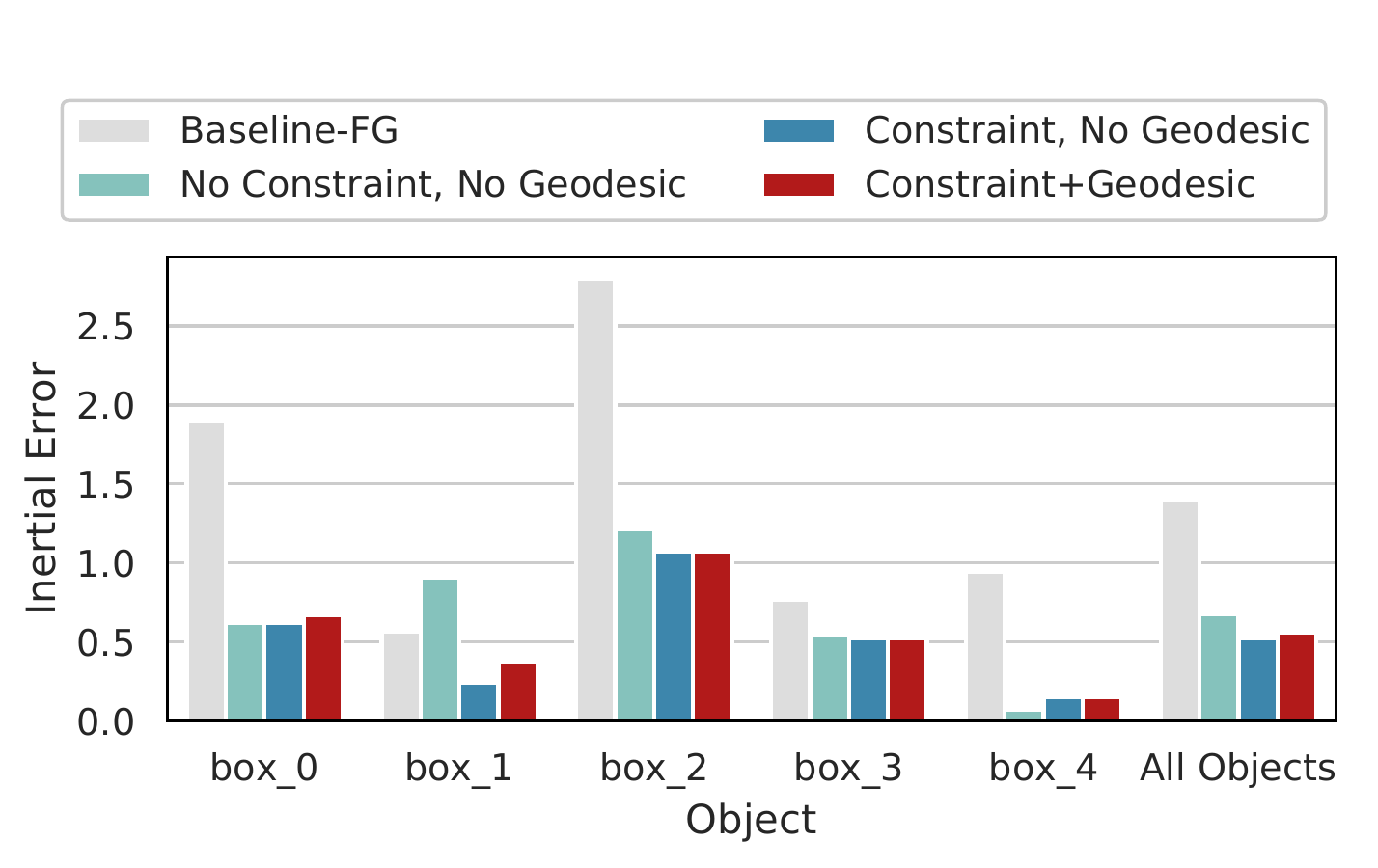}\\
  \includegraphics[width=0.49\textwidth]{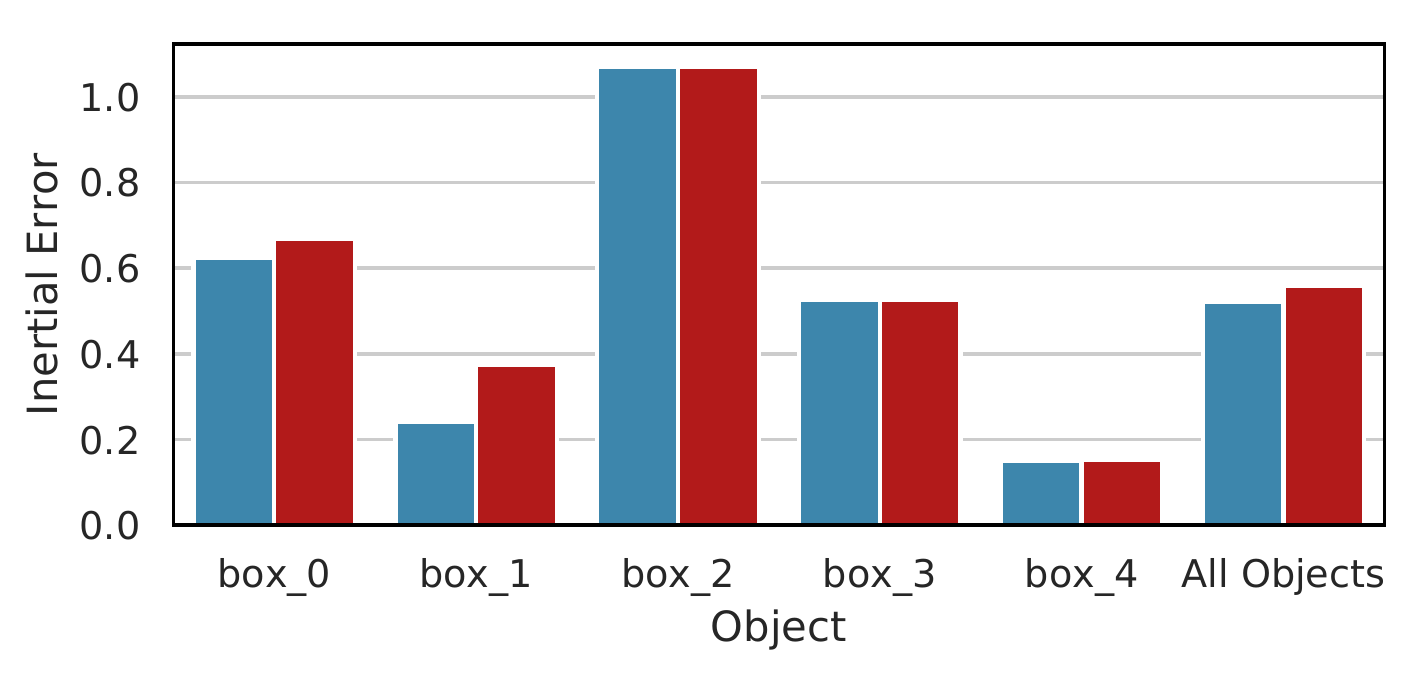}\\
  \caption{The inertial error is shown for the objects when mounted to the wrist force torque sensor~(``Wrist-FT'') \HL{in the real world}. We report all methods in the top plot and show only our two best methods in the bottom. We see that all methods except~``Baseline-FG'' estimates the inertial parameters sufficiently well.}
\label{fig:ft_inertial}
\end{figure}

By attaching the object directly to a wrist force torque sensor, the sensed forces and torques are only due to the object's dynamics, enabling accurate estimation of the inertial parameters. We see that parameterizing the inertial parameters over the~$\mathbf{SO}(3)$ sufficiently recovers the inertial parameters as shown by the performance of ``No Constraint, No Geodesic'' method in Fig.~\ref{fig:ft_inertial}. Adding inequality constraints for physical consistency~(``Constraint, No Geodesic'') performs better overall compared to just using the~$\mathbf{SO}(3)$ manifold.

\subsubsection{In-Hand}
\label{sec:inert-estim-hand}
\begin{figure}[t]
  \centering
  \includegraphics[width=0.484\textwidth,right]{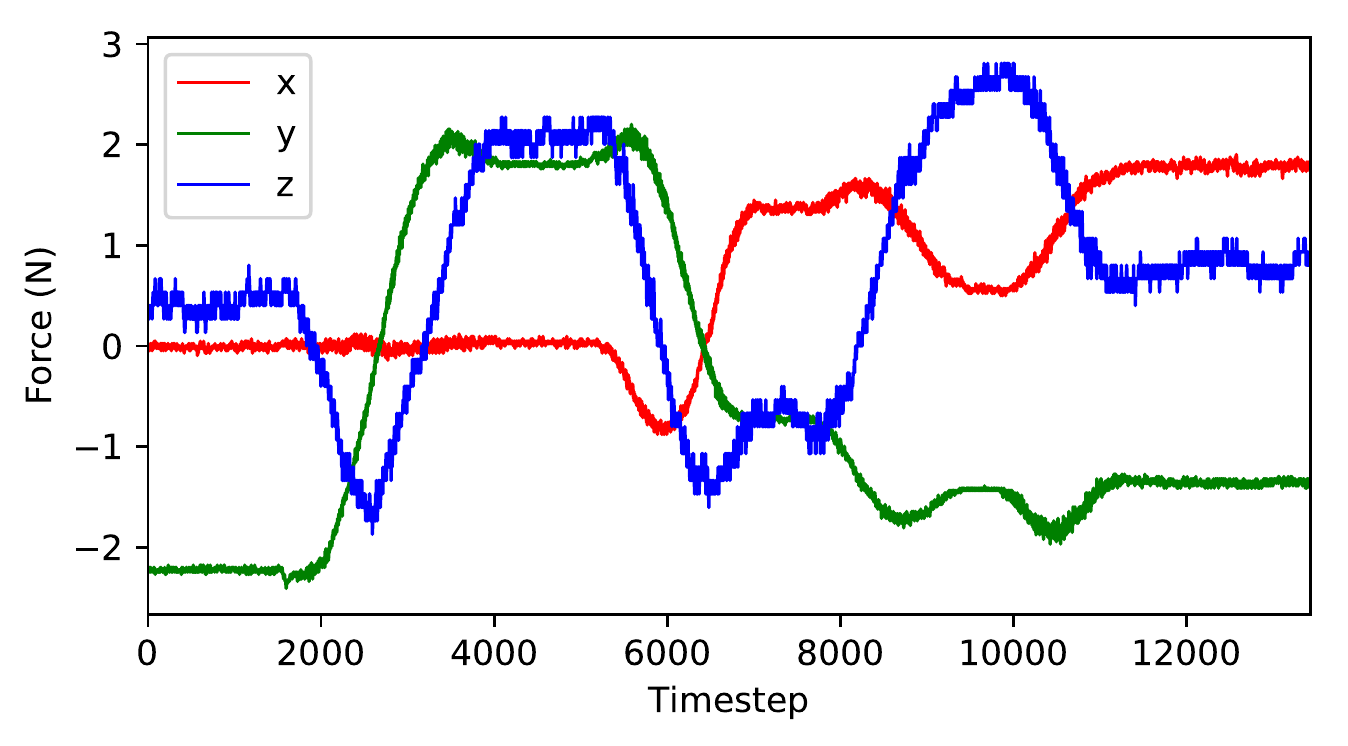}\\
  \includegraphics[width=0.484\textwidth,right]{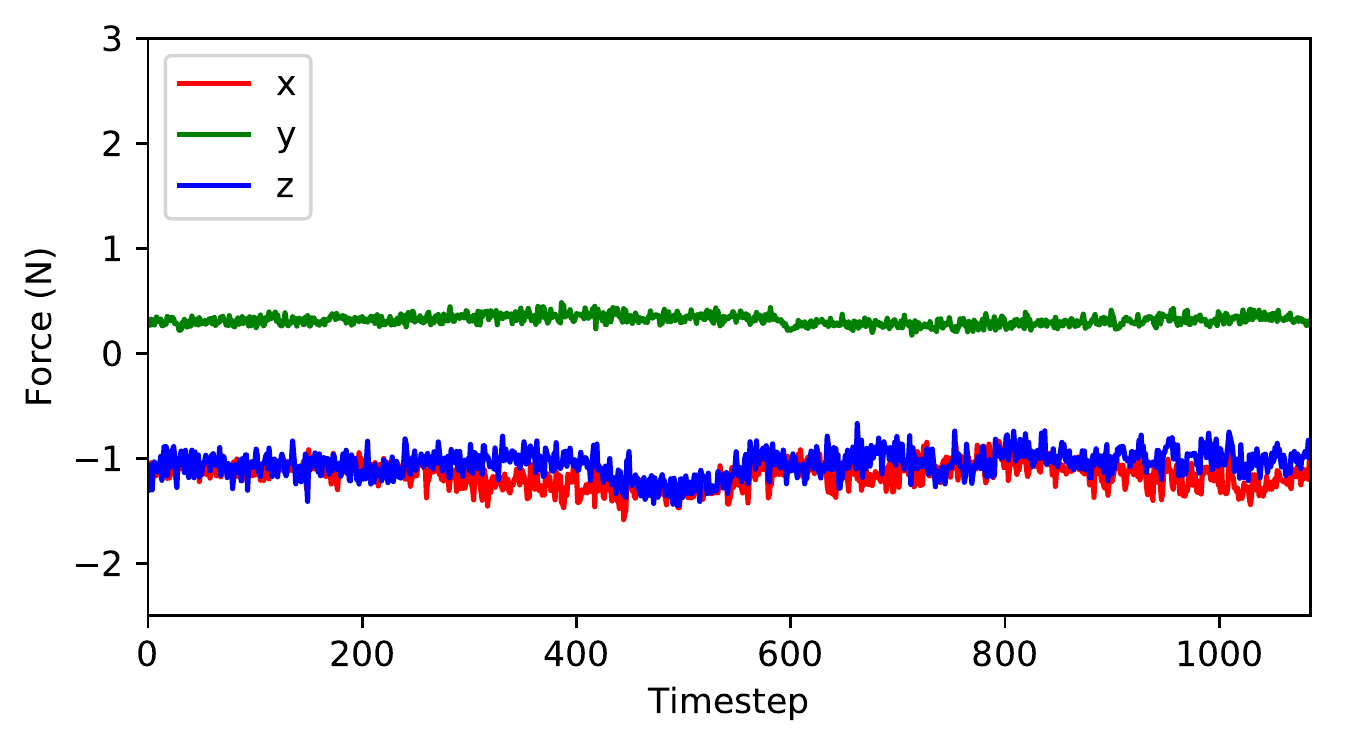}
  \caption{The sensed forces from the ``Wrist-FT'' data source is shown in top and \HL{the net force estimated by the BioTacs for ``In-Hand'' data source is shown with respect to the object's frame in the bottom for~box\_2 object.} We see that ``In-Hand'' force changes are very small compared to ``Wrist-FT''. \HL{Both plots use data from the real world.}}
\label{fig:force_traj}
\end{figure}

\begin{figure}[t]
  \centering \includegraphics[trim={0 0cm 0 1cm},clip,width=0.49\textwidth]{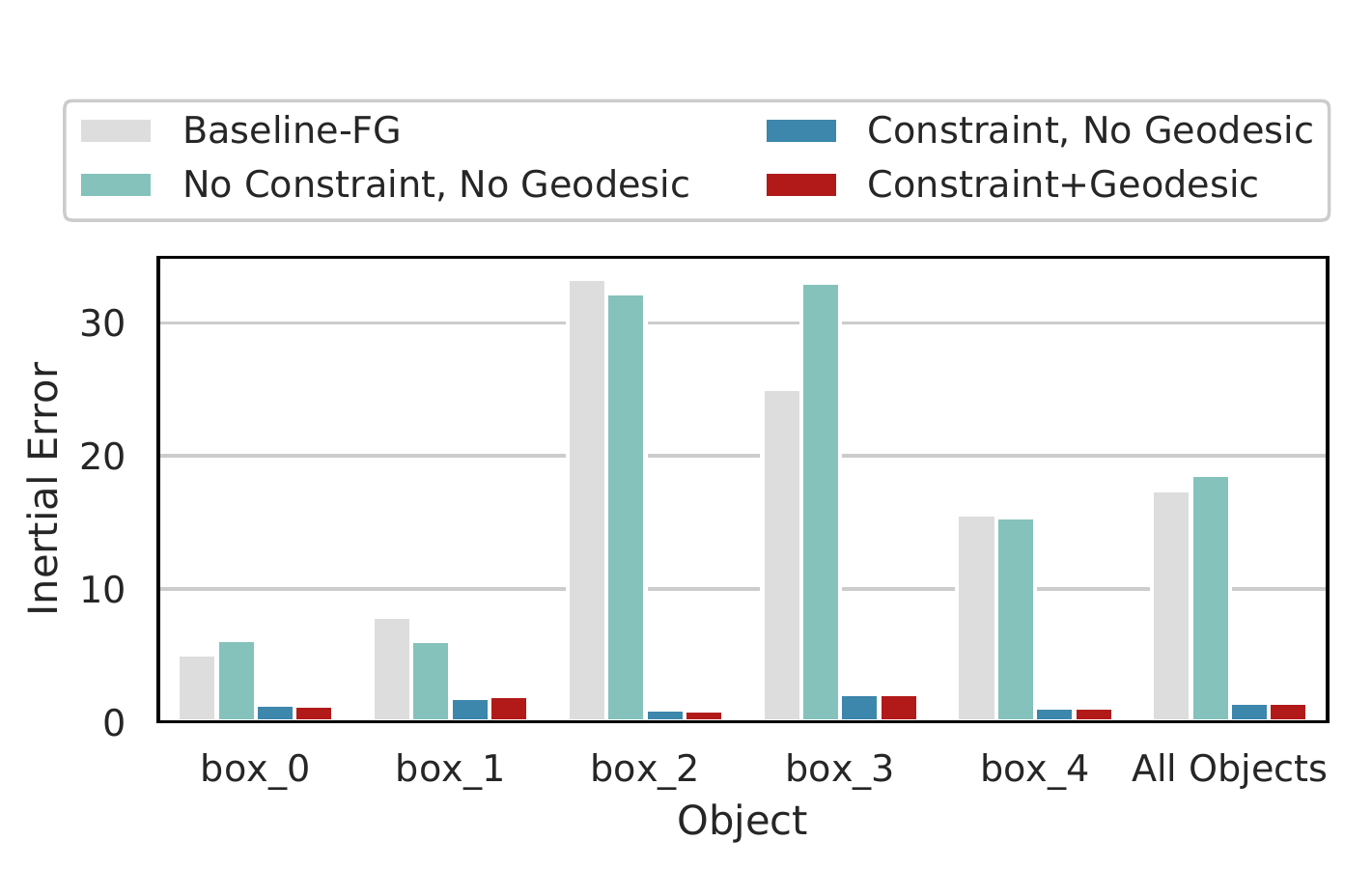}\\
  \includegraphics[width=0.49\textwidth]{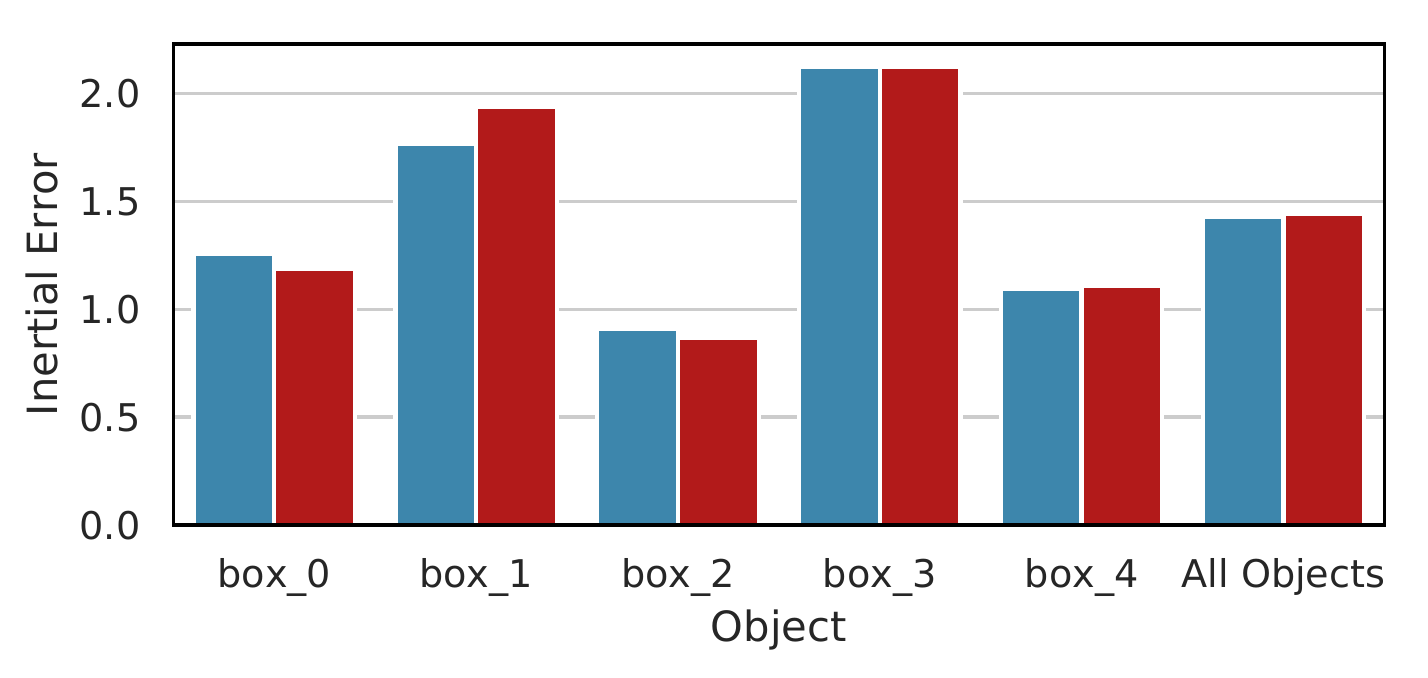}\\
  \caption{The inertial error during \HL{real-world} in-hand motion~\HL{(``In-Hand'')} is shown for objects when tracked using vision. We show all methods in the top plot and our two best methods are shown in the bottom plot. \HL{The force for inertial estimation is estimated from the BioTacs using our multi-sample dropout method.}}
\label{fig:dart_bt}
\end{figure}

\begin{figure}[t]
  \centering
  \includegraphics[width=0.49\textwidth,trim={0 0cm 0 1cm},clip]{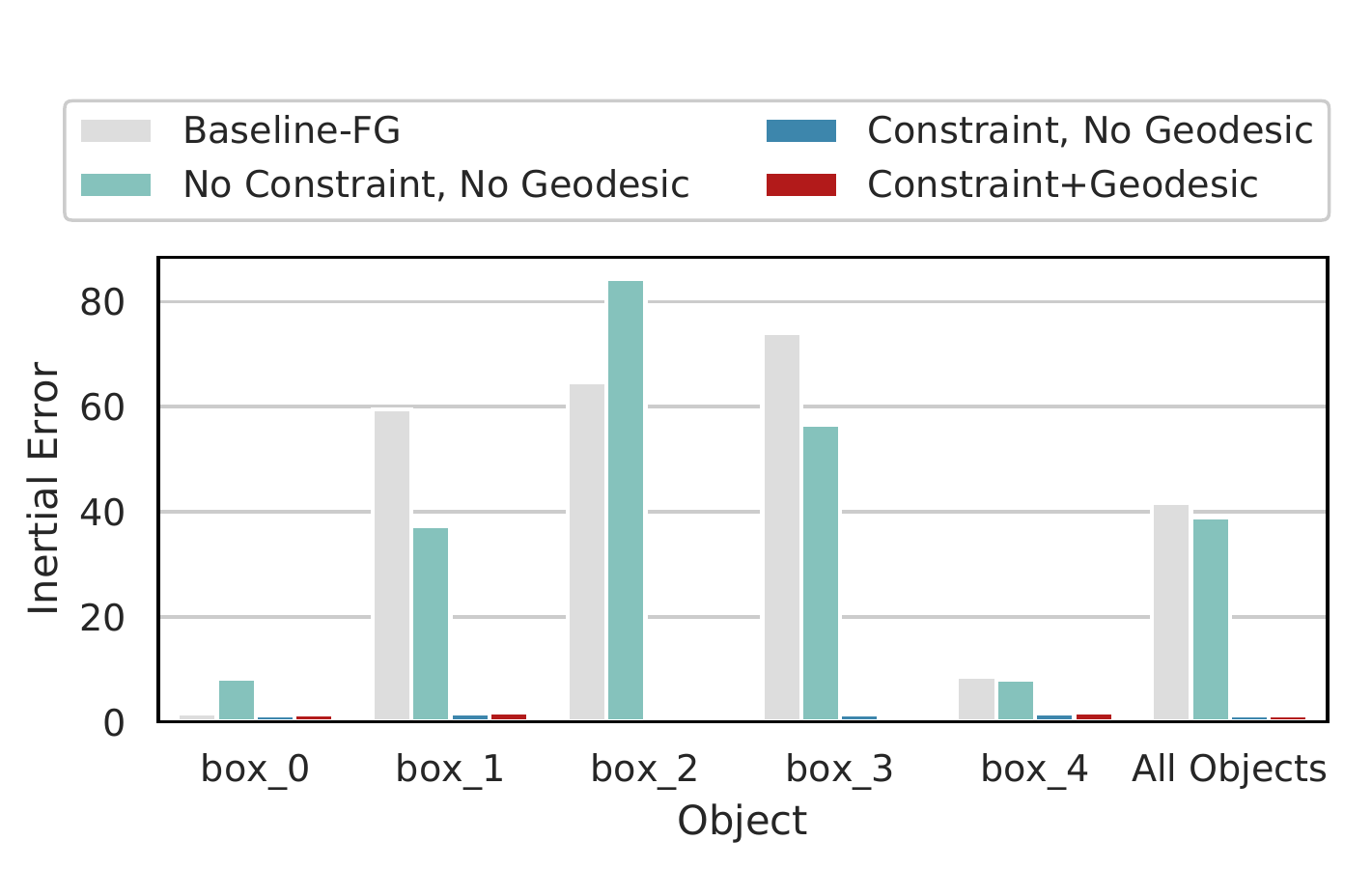}\\
  \includegraphics[width=0.49\textwidth]{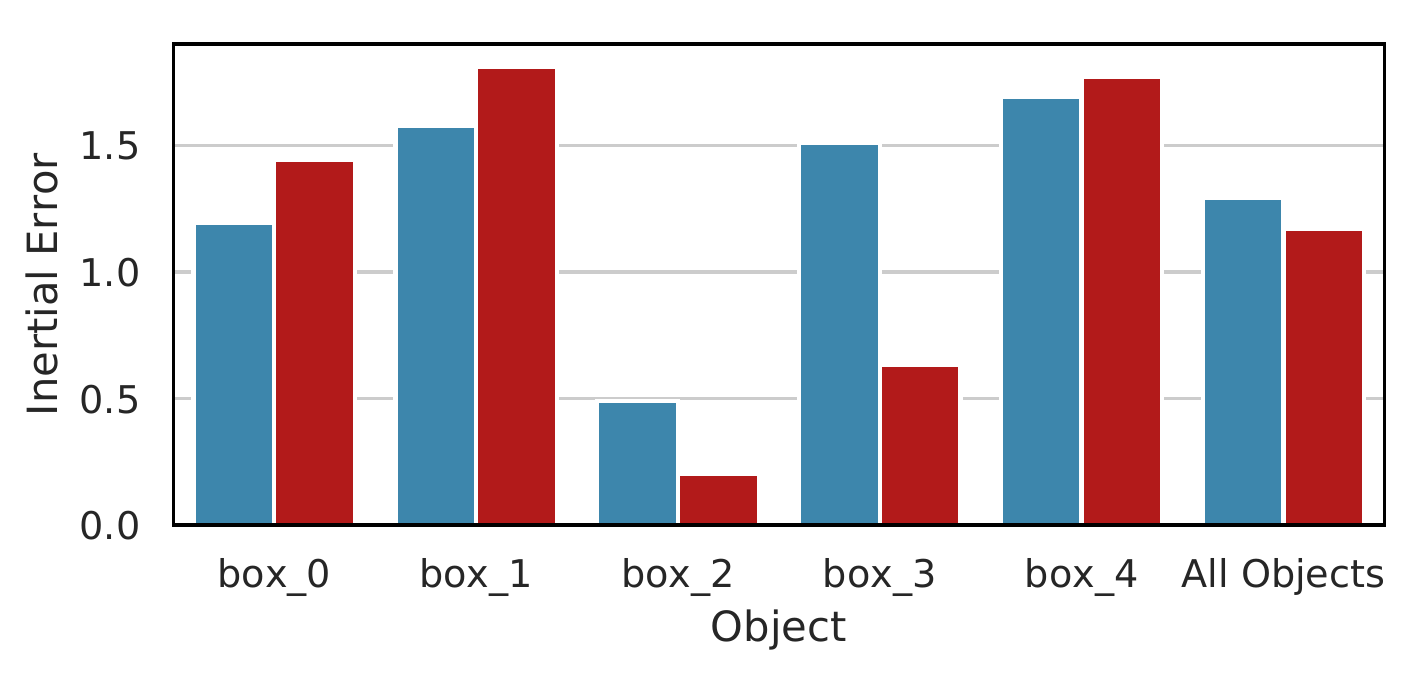}\\
  \caption{The inertial error during \HL{real-world} in-hand motion~\HL(``In-Hand'') is shown for the objects when tracked kinematically using ``relaxed-rigidity'' constraints. All methods are shown in the top plot and our two best methods are shown in the bottom plot. \HL{The force for inertial estimation is estimated from the BioTacs using our multi-sample dropout method.}}
\label{fig:rr_bt}
\end{figure}

When holding the object in-hand, the sensed force at one fingertip contains both the force due to the object dynamics and the force applied by the other fingertips transmitted through the object. Hence changes in the sensed force due to object dynamics excitation will be at a smaller scale than when the object is rigidly attached to just one sensor as seen in Fig.~\ref{fig:force_traj}. We are able to estimate the object's dynamics under these challenges sufficiently well as shown by Fig.~\ref{fig:dart_bt} and Fig.~\ref{fig:rr_bt}. From the plots, it is clear that adding physical consistency in the optimization enables inferring parameters that are close to the ground truth object dynamics. Using ``relaxed-rigidity'' constraints to track the object shows promise in getting close to the true inertial parameters, better than using a vision based tracker. For~``box\_4'', \HL{we observed} the ring finger broke and made contact multiple times during object motion. We still recover the inertial parameters as our structured approach naturally accounts for contact changes~(e.g., when contacts break, the sensed force is zero). 

\HL{Looking at ``box\_2'' results, we found ``In-Hand'' to estimate the inertial parameters better than ``Wrist-FT''. We suspect this to be because of the contact pattern between the tactile fingertips and the object. The contact pattern helps generate better observation of the forces during dynamics excitation of~``box\_2''. We attribute this to the center of mass of ``box\_2'' being closer to the centroid of the contact pattern while it's further away for other objects. The effect of contact pattern on the object dynamics estimation requires further investigation and will be part of future work.} Across all our data sources, we found that the geodesic prior helped only in a small way compared to the significant improvement we found with adding inequality constraints for physical  consistency. We suspect this to be because of the prior having a small weight compared to the other factors.

%%%Local Variables:
%%% mode: latex
%%% TeX-master: "../paperdraft"
%%% End:

\section{Discussion}
\label{sec:discussion}
We highlight some of the key challenges existing in inferring object dynamics in-hand in this section and discuss some future directions to improve our current approach.

Using biologically inspired tactile sensors enables compliant interactions but limits the accuracy of sensing forces to learned models~\cite{su2015force,Sundaralingam-ICRA-19}. While learned models show promising accuracy, transient effects due to many moving components in the sensor might require additional study. This is especially important when moving objects in-hand at high speeds. Current tactile sensors also have only a small region of sensing compared to their whole geometry. With the BioTacs, the mounting frame often makes contact with the object and this prevents the sensor from observing the correct force applied by the finger on the object. The unobservable residual forces from each fingertip reduce the robot's ability to recover the object's dynamics. We plan on developing tactile sensors for fingertips with large sensing regions and unobtrusive mounting frames to alleviate this issue.

Inference over factor graphs enable a structured approach to state and parameter estimation. However, implementing inequality constraints have been limited to toggling the constraint as a cost if its violated. In this paper, we observed that constraint as a cost is sufficient to push the parameters towards the feasible region but not enough to ensure feasibility. We hope to incorporate efficient sparse nonlinear constrained optimization solvers~\cite{gill2005} to perform inference over the factor graph to ensure feasibility in the future.

Our current formulation of the inertial inference does not include control input or the robot's forward dynamics model. This is because, the hand that we used does not have accurate torque feedback. We hope to adapt recent work on formulating robot dynamics as a factor graph~\cite{m2019unified} to study this problem in the future. \HL{Additionally, our method does not account for changing mass or changing mass distribution which is common in objects that have articulations or containers with filled liquids. We also do not account for the dynamics existent in deformable objects. We hope to explore these limitations in the future.}

Previously, quasi-static assumptions have enabled smoothing by enforcing constant velocity~\cite{lambert-2019icra-vistac}. This assumption fails for our problem as our dynamics allows impulse response in the system. Exploring smoothing over dynamical systems is part of our future work. For robots in unstructured environments, it would be helpful if they could autonomously interact with the unknown objects to acquire knowledge. To perform autonomous interaction, methods that do not rely on object dynamics could be leveraged to perform small manipulations of the object~\cite{lu-ram2020-grasp-inference,sundaralingam2019relaxed}. The extensive work in observability aware estimation  from other domains~\cite{calafiore2001robot,armstrong1989finding,caccavale1994identification,gautier1992exciting,van1994practical,Preiss2018,albee2019combining,Wilson2015} could be leveraged to actively interact with the object to improve estimation.

%%% Local Variables:
%%% mode: latex
%%% TeX-master: "../paperdraft"
%%% End:

%\clearpage

\section{Conclusion}
\label{sec:conclusion}
This paper shows how tactile fingertips can be leveraged to estimate object dynamics. Specifically, we formulate the in-hand dynamics between fingertips and the grasped object in a factor graph, encoding observations from different sensors. We also formulate the inertial parameters to be physically consistent to enable inertial estimation under partial excitation of the object's dynamics. Our friction estimation also enables accurate friction coefficient estimation on a varied set of objects.
\section*{Acknowledgments}
We would like to thank Yashraj Narang for discussions on ground truth computation and Ankur Handa for help with multi-sample dropout.  Balakumar Sundaralingam was supported by NSF Awards \#1657596 and
\#1846341.

\bibliography{ref_sysid_v2}
\onecolumn
\begin{appendices}
\section{Numerical results}
\label{ap:tables}
We report the numerical results in the tables below. We highlight the ground truth values with a yellow background and our approach without requiring an initial inertial estimate with a green background.

\begin{table}[h]
  \caption{Estimated inertial parameters from our simulation data source.}
  \label{tab:inertial-sim}
  \centering
\begin{tabular}{|l|l|r|r|r|r|r|r|r|r|r|r|}
\hline
\multicolumn{1}{|c|}{}                                  & \multicolumn{1}{c|}{}                                  & \multicolumn{1}{c|}{}                                     & \multicolumn{3}{c|}{\textbf{Center of Mass (m)}}                                                    & \multicolumn{6}{c|}{\textbf{Inertia ($10^2$ kg $m^2$) }}                                                                                                                                                         \\ \cline{4-12} 
\multicolumn{1}{|c|}{\multirow{-2}{*}{\textbf{Object}}} & \multicolumn{1}{c|}{\multirow{-2}{*}{\textbf{Method}}} & \multicolumn{1}{c|}{\multirow{-2}{*}{\textbf{mass (kg)}}} & \multicolumn{1}{c|}{\textbf{x}} & \multicolumn{1}{c|}{\textbf{y}} & \multicolumn{1}{c|}{\textbf{z}} & \multicolumn{1}{c|}{\textbf{xx}} & \multicolumn{1}{c|}{\textbf{yy}} & \multicolumn{1}{c|}{\textbf{zz}} & \multicolumn{1}{c|}{\textbf{xy}} & \multicolumn{1}{c|}{\textbf{xz}} & \multicolumn{1}{c|}{\textbf{yz}} \\ \hline
                                                        & \cellcolor[HTML]{FFFC9E}GT                             & \cellcolor[HTML]{FFFC9E}1.300                             & \cellcolor[HTML]{FFFC9E}0.200   & \cellcolor[HTML]{FFFC9E}0.500   & \cellcolor[HTML]{FFFC9E}0.100   & \cellcolor[HTML]{FFFC9E}40.15    & \cellcolor[HTML]{FFFC9E}12.92    & \cellcolor[HTML]{FFFC9E}47.99    & \cellcolor[HTML]{FFFC9E}-12.97   & \cellcolor[HTML]{FFFC9E}-2.59    & \cellcolor[HTML]{FFFC9E}-6.49    \\ \cline{2-12} 
                                                        & Baseline                                               & 0.073                                                     & 0.086                           & 0.419                           & 0.068                           & 1.76                             & -0.79                            & 1.88                             & 0.15                             & -0.29                            & -0.29                            \\ \cline{2-12} 
                                                        & Baseline-FG                                                 & 1.292                                                     & 0.199                           & 0.488                           & 0.097                           & 22.80                            & 5.38                             & 35.80                            & -9.00                            & -2.90                            & \cellcolor[HTML]{FFFFFF}-6.06    \\ \cline{2-12} 
                                                        & No C, No G                                             & 1.292                                                     & 0.198                           & 0.486                           & 0.095                           & 14.17                            & -2.71                            & 34.89                            & -0.20                            & -2.61                            & -5.28                            \\ \cline{2-12} 
                                                        & \cellcolor[HTML]{9AFF99}C, No G                        & \cellcolor[HTML]{9AFF99}1.291                             & \cellcolor[HTML]{9AFF99}0.197   & \cellcolor[HTML]{9AFF99}0.486   & \cellcolor[HTML]{9AFF99}0.095   & \cellcolor[HTML]{9AFF99}18.64    & \cellcolor[HTML]{9AFF99}17.21    & \cellcolor[HTML]{9AFF99}36.26    & \cellcolor[HTML]{9AFF99}0.14     & \cellcolor[HTML]{9AFF99}-0.93    & \cellcolor[HTML]{9AFF99}-2.66    \\ \cline{2-12} 
\multirow{-6}{*}{Sim ($\sigma^2=0$)}                    & C + G                                                  & 1.291                                                     & 0.197                           & 0.486                           & 0.095                           & 18.64                            & 17.21                            & 36.26                            & 0.14                             & -0.93                            & -2.66                            \\ \hline
                                                        & \cellcolor[HTML]{FFFC9E}GT                             & \cellcolor[HTML]{FFFC9E}1.300                             & \cellcolor[HTML]{FFFC9E}0.200   & \cellcolor[HTML]{FFFC9E}0.500   & \cellcolor[HTML]{FFFC9E}0.100   & \cellcolor[HTML]{FFFC9E}40.15    & \cellcolor[HTML]{FFFC9E}12.92    & \cellcolor[HTML]{FFFC9E}47.99    & \cellcolor[HTML]{FFFC9E}-12.97   & \cellcolor[HTML]{FFFC9E}-2.59    & \cellcolor[HTML]{FFFC9E}-6.49    \\ \cline{2-12} 
                                                        & Baseline                                               & 0.076                                                     & 0.077                           & 0.422                           & 0.048                           & 1.83                             & -1.23                            & 1.90                             & -0.15                            & -0.21                            & -0.21                            \\ \cline{2-12} 
                                                        & Baseline-FG                                                 & 1.292                                                     & 0.199                           & 0.486                           & 0.097                           & 24.38                            & 4.71                             & 35.46                            & -8.95                            & -2.18                            & -6.17                            \\ \cline{2-12} 
                                                        & No C, No G                                             & 1.294                                                     & 0.198                           & 0.485                           & 0.095                           & 10.41                            & -1.93                            & 30.99                            & 0.05                             & -1.80                            & -4.28                            \\ \cline{2-12} 
                                                        & \cellcolor[HTML]{9AFF99}C, No G                        & \cellcolor[HTML]{9AFF99}1.293                             & \cellcolor[HTML]{9AFF99}0.197   & \cellcolor[HTML]{9AFF99}0.485   & \cellcolor[HTML]{9AFF99}0.094   & \cellcolor[HTML]{9AFF99}18.61    & \cellcolor[HTML]{9AFF99}17.32    & \cellcolor[HTML]{9AFF99}36.04    & \cellcolor[HTML]{9AFF99}0.09     & \cellcolor[HTML]{9AFF99}-0.70    & \cellcolor[HTML]{9AFF99}-2.41    \\ \cline{2-12} 
\multirow{-6}{*}{Sim ($\sigma^2=0.1$)}                  & C + G                                                  & 1.293                                                     & 0.197                           & 0.485                           & 0.094                           & 18.61                            & 17.32                            & 36.04                            & 0.09                             & -0.70                            & -2.41                            \\ \hline
                                                        & \cellcolor[HTML]{FFFC9E}GT                             & \cellcolor[HTML]{FFFC9E}1.300                             & \cellcolor[HTML]{FFFC9E}0.200   & \cellcolor[HTML]{FFFC9E}0.500   & \cellcolor[HTML]{FFFC9E}0.100   & \cellcolor[HTML]{FFFC9E}40.15    & \cellcolor[HTML]{FFFC9E}12.92    & \cellcolor[HTML]{FFFC9E}47.99    & \cellcolor[HTML]{FFFC9E}-12.97   & \cellcolor[HTML]{FFFC9E}-2.59    & \cellcolor[HTML]{FFFC9E}-6.49    \\ \cline{2-12} 
                                                        & Baseline                                               & 0.076                                                     & 0.073                           & 0.355                           & 0.074                           & 1.29                             & -1.17                            & 1.80                             & 1.27                             & -0.34                            & -0.55                            \\ \cline{2-12} 
                                                        & Baseline-FG                                                 & 1.289                                                     & 0.195                           & 0.485                           & 0.098                           & 22.16                            & 4.44                             & 34.76                            & -8.24                            & -1.54                            & -6.29                            \\ \cline{2-12} 
                                                        & No C, No G                                             & 1.288                                                     & 0.196                           & 0.487                           & 0.099                           & 10.56                            & -1.98                            & 27.37                            & -0.02                            & -1.24                            & -3.86                            \\ \cline{2-12} 
                                                        & \cellcolor[HTML]{9AFF99}C, No G                        & \cellcolor[HTML]{9AFF99}1.288                             & \cellcolor[HTML]{9AFF99}0.196   & \cellcolor[HTML]{9AFF99}0.487   & \cellcolor[HTML]{9AFF99}0.097   & \cellcolor[HTML]{9AFF99}18.13    & \cellcolor[HTML]{9AFF99}19.18    & \cellcolor[HTML]{9AFF99}35.74    & \cellcolor[HTML]{9AFF99}0.08     & \cellcolor[HTML]{9AFF99}-0.57    & \cellcolor[HTML]{9AFF99}-2.36    \\ \cline{2-12} 
\multirow{-6}{*}{Sim ($\sigma^2=0.25$)}                 & C + G                                                  & 1.288                                                     & 0.196                           & 0.487                           & 0.097                           & 18.13                            & 19.18                            & 35.74                            & 0.08                             & -0.57                            & -2.36                            \\ \hline
                                                        & \cellcolor[HTML]{FFFC9E}GT                             & \cellcolor[HTML]{FFFC9E}1.300                             & \cellcolor[HTML]{FFFC9E}0.200   & \cellcolor[HTML]{FFFC9E}0.500   & \cellcolor[HTML]{FFFC9E}0.100   & \cellcolor[HTML]{FFFC9E}40.15    & \cellcolor[HTML]{FFFC9E}12.92    & \cellcolor[HTML]{FFFC9E}47.99    & \cellcolor[HTML]{FFFC9E}-12.97   & \cellcolor[HTML]{FFFC9E}-2.59    & \cellcolor[HTML]{FFFC9E}-6.49    \\ \cline{2-12} 
                                                        & Baseline                                               & 0.069                                                     & 0.102                           & 0.386                           & 0.115                           & 1.41                             & -0.25                            & 1.91                             & 0.34                             & -0.33                            & -0.42                            \\ \cline{2-12} 
                                                        & Baseline-FG                                                 & 1.293                                                     & 0.196                           & 0.481                           & 0.098                           & 23.24                            & 3.65                             & 34.27                            & -7.78                            & -2.37                            & -5.83                            \\ \cline{2-12} 
                                                        & No C, No G                                             & 1.296                                                     & 0.193                           & 0.481                           & 0.096                           & 10.28                            & -2.46                            & 23.69                            & -0.17                            & -1.12                            & -2.32                            \\ \cline{2-12} 
                                                        & \cellcolor[HTML]{9AFF99}C, No G                        & \cellcolor[HTML]{9AFF99}1.294                             & \cellcolor[HTML]{9AFF99}0.194   & \cellcolor[HTML]{9AFF99}0.481   & \cellcolor[HTML]{9AFF99}0.092   & \cellcolor[HTML]{9AFF99}15.05    & \cellcolor[HTML]{9AFF99}20.51    & \cellcolor[HTML]{9AFF99}34.95    & \cellcolor[HTML]{9AFF99}0.12     & \cellcolor[HTML]{9AFF99}-1.17    & \cellcolor[HTML]{9AFF99}-1.86    \\ \cline{2-12} 
\multirow{-6}{*}{Sim ($\sigma^2=0.5$)}                  & C + G                                                  & 1.294                                                     & 0.194                           & 0.481                           & 0.092                           & 15.05                            & 20.51                            & 34.95                            & 0.12                             & -1.17                            & -1.86                            \\ \hline
                                                        & \cellcolor[HTML]{FFFC9E}GT                             & \cellcolor[HTML]{FFFC9E}1.300                             & \cellcolor[HTML]{FFFC9E}0.200   & \cellcolor[HTML]{FFFC9E}0.500   & \cellcolor[HTML]{FFFC9E}0.100   & \cellcolor[HTML]{FFFC9E}40.15    & \cellcolor[HTML]{FFFC9E}12.92    & \cellcolor[HTML]{FFFC9E}47.99    & \cellcolor[HTML]{FFFC9E}-12.97   & \cellcolor[HTML]{FFFC9E}-2.59    & \cellcolor[HTML]{FFFC9E}-6.49    \\ \cline{2-12} 
                                                        & Baseline                                               & 0.068                                                     & 0.159                           & 0.492                           & 0.073                           & 1.14                             & 0.11                             & 2.09                             & -0.33                            & -0.05                            & -0.22                            \\ \cline{2-12} 
                                                        & Baseline-FG                                                 & 1.277                                                     & 0.194                           & 0.478                           & 0.107                           & 24.41                            & 4.29                             & 33.02                            & -8.43                            & -1.97                            & -6.48                            \\ \cline{2-12} 
                                                        & No C, No G                                             & 1.269                                                     & 0.196                           & 0.485                           & 0.104                           & 8.61                             & -1.76                            & 24.21                            & -0.06                            & -0.91                            & -3.14                            \\ \cline{2-12} 
                                                        & \cellcolor[HTML]{9AFF99}C, No G                        & \cellcolor[HTML]{9AFF99}1.273                             & \cellcolor[HTML]{9AFF99}0.196   & \cellcolor[HTML]{9AFF99}0.480   & \cellcolor[HTML]{9AFF99}0.099   & \cellcolor[HTML]{9AFF99}13.93    & \cellcolor[HTML]{9AFF99}20.07    & \cellcolor[HTML]{9AFF99}33.71    & \cellcolor[HTML]{9AFF99}0.05     & \cellcolor[HTML]{9AFF99}-0.43    & \cellcolor[HTML]{9AFF99}-1.37    \\ \cline{2-12} 
\multirow{-6}{*}{Sim ($\sigma^2=1$)}                    & C + G                                                  & 1.273                                                     & 0.196                           & 0.480                           & 0.099                           & 13.93                            & 20.07                            & 33.71                            & 0.05                             & -0.43                            & -1.37                            \\ \hline
\end{tabular}
\end{table}
\begin{table}[h]
  \centering
  \caption{Estimated inertial parameters from ``Wrist-FT'' data source.}
  \label{tab:inertial-ft}

\begin{tabular}{|l|l|r|r|r|r|r|r|r|r|r|r|}
\hline
\multicolumn{1}{|c|}{}                                  & \multicolumn{1}{c|}{}                                  & \multicolumn{1}{c|}{}                                     & \multicolumn{3}{c|}{\textbf{Center of Mass (m)}}                                                    & \multicolumn{6}{c|}{\textbf{Inertia ($10^5$ kg $m^2$)}}                                                                                                                                                         \\ \cline{4-12} 
\multicolumn{1}{|c|}{\multirow{-2}{*}{\textbf{Object}}} & \multicolumn{1}{c|}{\multirow{-2}{*}{\textbf{Method}}} & \multicolumn{1}{c|}{\multirow{-2}{*}{\textbf{mass (kg)}}} & \multicolumn{1}{c|}{\textbf{x}} & \multicolumn{1}{c|}{\textbf{y}} & \multicolumn{1}{c|}{\textbf{z}} & \multicolumn{1}{c|}{\textbf{xx}} & \multicolumn{1}{c|}{\textbf{yy}} & \multicolumn{1}{c|}{\textbf{zz}} & \multicolumn{1}{c|}{\textbf{xy}} & \multicolumn{1}{c|}{\textbf{xz}} & \multicolumn{1}{c|}{\textbf{yz}} \\ \hline
                                                        & \cellcolor[HTML]{FFFC9E}GT                             & \cellcolor[HTML]{FFFC9E}0.115                             & \cellcolor[HTML]{FFFC9E}0.094   & \cellcolor[HTML]{FFFC9E}-0.037  & \cellcolor[HTML]{FFFC9E}0.032   & \cellcolor[HTML]{FFFC9E}37.20    & \cellcolor[HTML]{FFFC9E}150.80   & \cellcolor[HTML]{FFFC9E}154.00   & \cellcolor[HTML]{FFFC9E}-40.00   & \cellcolor[HTML]{FFFC9E}34.80    & \cellcolor[HTML]{FFFC9E}-13.60   \\ \cline{2-12} 
                                                        & Baseline-FG                                                 & 0.147                                                     & 0.072                           & -0.022                          & 0.067                           & 85.29                            & 87.63                            & 47.73                            & -61.94                           & 12.39                            & \cellcolor[HTML]{FFFFFF}-52.78   \\ \cline{2-12} 
                                                        & No C, No G                                             & 0.148                                                     & 0.057                           & -0.010                          & 0.037                           & 82.88                            & 135.25                           & 59.30                            & -49.09                           & 23.35                            & -29.76                           \\ \cline{2-12} 
                                                        & \cellcolor[HTML]{9AFF99}C, No G                        & \cellcolor[HTML]{9AFF99}0.148                             & \cellcolor[HTML]{9AFF99}0.057   & \cellcolor[HTML]{9AFF99}-0.010  & \cellcolor[HTML]{9AFF99}0.037   & \cellcolor[HTML]{9AFF99}82.88    & \cellcolor[HTML]{9AFF99}135.25   & \cellcolor[HTML]{9AFF99}59.30    & \cellcolor[HTML]{9AFF99}-49.09   & \cellcolor[HTML]{9AFF99}23.35    & \cellcolor[HTML]{9AFF99}-29.76   \\ \cline{2-12} 
\multirow{-5}{*}{box\_0}                                & C + G                                                  & 0.148                                                     & 0.057                           & -0.010                          & 0.038                           & 82.45                            & 144.93                           & 59.00                            & -49.08                           & 23.34                            & -29.76                           \\ \hline
                                                        & \cellcolor[HTML]{FFFC9E}GT                             & \cellcolor[HTML]{FFFC9E}0.179                             & \cellcolor[HTML]{FFFC9E}0.071   & \cellcolor[HTML]{FFFC9E}-0.033  & \cellcolor[HTML]{FFFC9E}0.031   & \cellcolor[HTML]{FFFC9E}50.00    & \cellcolor[HTML]{FFFC9E}165.30   & \cellcolor[HTML]{FFFC9E}166.90   & \cellcolor[HTML]{FFFC9E}-45.10   & \cellcolor[HTML]{FFFC9E}40.70    & \cellcolor[HTML]{FFFC9E}-18.80   \\ \cline{2-12} 
                                                        & Baseline-FG                                                 & 0.195                                                     & 0.059                           & -0.013                          & 0.027                           & 105.63                           & 428.65                           & 74.85                            & -53.87                           & 32.51                            & -11.93                           \\ \cline{2-12} 
                                                        & No C, No G                                             & 0.195                                                     & 0.056                           & -0.012                          & 0.030                           & 106.00                           & 357.68                           & 73.92                            & -64.52                           & 30.81                            & -37.92                           \\ \cline{2-12} 
                                                        & \cellcolor[HTML]{9AFF99}C, No G                        & \cellcolor[HTML]{9AFF99}0.195                             & \cellcolor[HTML]{9AFF99}0.051   & \cellcolor[HTML]{9AFF99}-0.011  & \cellcolor[HTML]{9AFF99}0.031   & \cellcolor[HTML]{9AFF99}106.41   & \cellcolor[HTML]{9AFF99}82.19    & \cellcolor[HTML]{9AFF99}74.31    & \cellcolor[HTML]{9AFF99}-64.56   & \cellcolor[HTML]{9AFF99}30.71    & \cellcolor[HTML]{9AFF99}-38.95   \\ \cline{2-12} 
\multirow{-5}{*}{box\_1}                                & C + G                                                  & 0.194                                                     & 0.060                           & -0.014                          & 0.030                           & 107.25                           & 85.89                            & 70.59                            & -64.31                           & 30.58                            & -38.93                           \\ \hline
                                                        & \cellcolor[HTML]{FFFC9E}GT                             & \cellcolor[HTML]{FFFC9E}0.179                             & \cellcolor[HTML]{FFFC9E}0.090   & \cellcolor[HTML]{FFFC9E}-0.033  & \cellcolor[HTML]{FFFC9E}0.031   & \cellcolor[HTML]{FFFC9E}50.00    & \cellcolor[HTML]{FFFC9E}204.60   & \cellcolor[HTML]{FFFC9E}206.20   & \cellcolor[HTML]{FFFC9E}-54.10   & \cellcolor[HTML]{FFFC9E}51.40    & \cellcolor[HTML]{FFFC9E}-18.80   \\ \cline{2-12} 
                                                        & Baseline-FG                                                 & 0.222                                                     & 0.071                           & 0.005                           & 0.062                           & 189.47                           & 123.73                           & 139.89                           & -103.96                          & 21.33                            & -93.52                           \\ \cline{2-12} 
                                                        & No C, No G                                             & 0.228                                                     & 0.051                           & 0.015                           & 0.032                           & 205.99                           & 94.38                            & 168.55                           & -75.20                           & 35.97                            & -47.08                           \\ \cline{2-12} 
                                                        & \cellcolor[HTML]{9AFF99}C, No G                        & \cellcolor[HTML]{9AFF99}0.215                             & \cellcolor[HTML]{9AFF99}0.071   & \cellcolor[HTML]{9AFF99}0.000   & \cellcolor[HTML]{9AFF99}0.033   & \cellcolor[HTML]{9AFF99}155.93   & \cellcolor[HTML]{9AFF99}195.55   & \cellcolor[HTML]{9AFF99}120.67   & \cellcolor[HTML]{9AFF99}-71.51   & \cellcolor[HTML]{9AFF99}34.01    & \cellcolor[HTML]{9AFF99}-43.31   \\ \cline{2-12} 
\multirow{-5}{*}{box\_2}                                & C + G                                                  & 0.215                                                     & 0.071                           & 0.000                           & 0.033                           & 155.93                           & 195.57                           & 120.67                           & -71.51                           & 34.01                            & -43.31                           \\ \hline
                                                        & \cellcolor[HTML]{FFFC9E}GT                             & \cellcolor[HTML]{FFFC9E}0.131                             & \cellcolor[HTML]{FFFC9E}0.099   & \cellcolor[HTML]{FFFC9E}-0.036  & \cellcolor[HTML]{FFFC9E}0.033   & \cellcolor[HTML]{FFFC9E}42.00    & \cellcolor[HTML]{FFFC9E}182.90   & \cellcolor[HTML]{FFFC9E}184.40   & \cellcolor[HTML]{FFFC9E}-45.80   & \cellcolor[HTML]{FFFC9E}44.60    & \cellcolor[HTML]{FFFC9E}-15.50   \\ \cline{2-12} 
                                                        & Baseline-FG                                                 & 0.192                                                     & 0.083                           & -0.056                          & 0.036                           & 51.99                            & 101.06                           & 39.46                            & -69.17                           & 27.34                            & -39.39                           \\ \cline{2-12} 
                                                        & No C, No G                                             & 0.171                                                     & 0.082                           & -0.023                          & 0.031                           & 79.86                            & 196.46                           & 63.65                            & -56.69                           & 26.98                            & -34.01                           \\ \cline{2-12} 
                                                        & \cellcolor[HTML]{9AFF99}C, No G                        & \cellcolor[HTML]{9AFF99}0.171                             & \cellcolor[HTML]{9AFF99}0.083   & \cellcolor[HTML]{9AFF99}-0.023  & \cellcolor[HTML]{9AFF99}0.031   & \cellcolor[HTML]{9AFF99}82.83    & \cellcolor[HTML]{9AFF99}158.30   & \cellcolor[HTML]{9AFF99}64.15    & \cellcolor[HTML]{9AFF99}-56.90   & \cellcolor[HTML]{9AFF99}27.05    & \cellcolor[HTML]{9AFF99}-33.85   \\ \cline{2-12} 
\multirow{-5}{*}{box\_3}                                & C + G                                                  & 0.171                                                     & 0.083                           & -0.023                          & 0.031                           & 82.82                            & 158.30                           & 64.14                            & -56.90                           & 27.05                            & -33.85                           \\ \hline
                                                        & \cellcolor[HTML]{FFFC9E}GT                             & \cellcolor[HTML]{FFFC9E}0.147                             & \cellcolor[HTML]{FFFC9E}0.109   & \cellcolor[HTML]{FFFC9E}-0.035  & \cellcolor[HTML]{FFFC9E}0.031   & \cellcolor[HTML]{FFFC9E}43.30    & \cellcolor[HTML]{FFFC9E}239.00   & \cellcolor[HTML]{FFFC9E}242.00   & \cellcolor[HTML]{FFFC9E}-53.70   & \cellcolor[HTML]{FFFC9E}49.60    & \cellcolor[HTML]{FFFC9E}-16.00   \\ \cline{2-12} 
                                                        & Baseline-FG                                                 & 0.213                                                     & 0.089                           & -0.053                          & 0.034                           & 58.85                            & 120.96                           & 54.32                            & -80.44                           & 30.83                            & -43.62                           \\ \cline{2-12} 
                                                        & No C, No G                                             & 0.180                                                     & 0.085                           & -0.014                          & 0.031                           & 75.44                            & 43.42                            & 85.34                            & -59.81                           & 28.46                            & -37.12                           \\ \cline{2-12} 
                                                        & \cellcolor[HTML]{9AFF99}C, No G                        & \cellcolor[HTML]{9AFF99}0.177                             & \cellcolor[HTML]{9AFF99}0.076   & \cellcolor[HTML]{9AFF99}-0.019  & \cellcolor[HTML]{9AFF99}0.031   & \cellcolor[HTML]{9AFF99}68.18    & \cellcolor[HTML]{9AFF99}50.46    & \cellcolor[HTML]{9AFF99}64.09    & \cellcolor[HTML]{9AFF99}-58.58   & \cellcolor[HTML]{9AFF99}27.91    & \cellcolor[HTML]{9AFF99}-35.49   \\ \cline{2-12} 
\multirow{-5}{*}{box\_4}                                & C + G                                                  & 0.177                                                     & 0.076                           & -0.019                          & 0.031                           & 67.96                            & 50.69                            & 64.15                            & -58.59                           & 27.91                            & -35.49                           \\ \hline
\end{tabular}
\end{table}
\begin{table}[h]
  \centering
  \caption{Estimated inertial parameters from our ``In-Hand'' data source while tracking object with vision.}
  \label{tab:inertial-est-dart}
% \usepackage{multirow}
\begin{tabular}{|l|l|r|r|r|r|r|r|r|r|r|r|}
\hline
\multicolumn{1}{|c|}{}                                  & \multicolumn{1}{c|}{}                                  & \multicolumn{1}{c|}{}                                     & \multicolumn{3}{c|}{\textbf{Center of Mass (m)}}                                                    & \multicolumn{6}{c|}{\textbf{Inertia ($10^5$ kg $m^2$) }}                                                                                                                                                         \\ \cline{4-12} 
\multicolumn{1}{|c|}{\multirow{-2}{*}{\textbf{Object}}} & \multicolumn{1}{c|}{\multirow{-2}{*}{\textbf{Method}}} & \multicolumn{1}{c|}{\multirow{-2}{*}{\textbf{mass (kg)}}} & \multicolumn{1}{c|}{\textbf{x}} & \multicolumn{1}{c|}{\textbf{y}} & \multicolumn{1}{c|}{\textbf{z}} & \multicolumn{1}{c|}{\textbf{xx}} & \multicolumn{1}{c|}{\textbf{yy}} & \multicolumn{1}{c|}{\textbf{zz}} & \multicolumn{1}{c|}{\textbf{xy}} & \multicolumn{1}{c|}{\textbf{xz}} & \multicolumn{1}{c|}{\textbf{yz}} \\ \hline
                                                        & \cellcolor[HTML]{FFFC9E}GT                             & \cellcolor[HTML]{FFFC9E}0.115                             & \cellcolor[HTML]{FFFC9E}0.094   & \cellcolor[HTML]{FFFC9E}-0.037  & \cellcolor[HTML]{FFFC9E}0.032   & \cellcolor[HTML]{FFFC9E}37.20    & \cellcolor[HTML]{FFFC9E}150.80   & \cellcolor[HTML]{FFFC9E}154.00   & \cellcolor[HTML]{FFFC9E}-40.00   & \cellcolor[HTML]{FFFC9E}34.80    & \cellcolor[HTML]{FFFC9E}-13.60   \\ \cline{2-12} 
                                                        & Baseline-FG                                                 & 0.124                                                     & 0.451                           & 0.335                           & -0.084                          & 1592.04                          & 3843.79                          & 6073.08                          & -2243.44                         & -23.59                           & \cellcolor[HTML]{FFFFFF}1988.83  \\ \cline{2-12} 
                                                        & No C, No G                                             & 0.090                                                     & 0.649                           & 0.281                           & -0.009                          & 834.25                           & 4923.81                          & 6242.05                          & -2011.14                         & -390.47                          & 1405.04                          \\ \cline{2-12} 
                                                        & \cellcolor[HTML]{9AFF99}C, No G                        & \cellcolor[HTML]{9AFF99}0.045                             & \cellcolor[HTML]{9AFF99}0.244   & \cellcolor[HTML]{9AFF99}-0.003  & \cellcolor[HTML]{9AFF99}0.018   & \cellcolor[HTML]{9AFF99}1.84     & \cellcolor[HTML]{9AFF99}251.99   & \cellcolor[HTML]{9AFF99}250.91   & \cellcolor[HTML]{9AFF99}3.63     & \cellcolor[HTML]{9AFF99}-18.65   & \cellcolor[HTML]{9AFF99}0.27     \\ \cline{2-12} 
\multirow{-5}{*}{box\_0}                                & C + G                                                  & 0.045                                                     & 0.249                           & -0.004                          & 0.019                           & 1.92                             & 259.16                           & 258.08                           & 4.81                             & -19.66                           & 0.36                             \\ \hline
                                                        & \cellcolor[HTML]{FFFC9E}GT                             & \cellcolor[HTML]{FFFC9E}0.179                             & \cellcolor[HTML]{FFFC9E}0.071   & \cellcolor[HTML]{FFFC9E}-0.033  & \cellcolor[HTML]{FFFC9E}0.031   & \cellcolor[HTML]{FFFC9E}50.00    & \cellcolor[HTML]{FFFC9E}165.30   & \cellcolor[HTML]{FFFC9E}166.90   & \cellcolor[HTML]{FFFC9E}-45.10   & \cellcolor[HTML]{FFFC9E}40.70    & \cellcolor[HTML]{FFFC9E}-18.80   \\ \cline{2-12} 
                                                        & Baseline-FG                                                 & 0.048                                                     & 0.469                           & -0.231                          & 0.125                           & 326.17                           & 1122.55                          & 1310.51                          & 512.78                           & -278.30                          & 136.04                           \\ \cline{2-12} 
                                                        & No C, No G                                             & 0.053                                                     & 0.391                           & -0.199                          & 0.105                           & 269.65                           & 877.33                           & 1032.11                          & 414.18                           & -219.14                          & 110.96                           \\ \cline{2-12} 
                                                        & \cellcolor[HTML]{9AFF99}C, No G                        & \cellcolor[HTML]{9AFF99}0.043                             & \cellcolor[HTML]{9AFF99}0.077   & \cellcolor[HTML]{9AFF99}-0.061  & \cellcolor[HTML]{9AFF99}0.006   & \cellcolor[HTML]{9AFF99}13.22    & \cellcolor[HTML]{9AFF99}9.59     & \cellcolor[HTML]{9AFF99}42.40    & \cellcolor[HTML]{9AFF99}3.22     & \cellcolor[HTML]{9AFF99}-1.01    & \cellcolor[HTML]{9AFF99}1.87     \\ \cline{2-12} 
\multirow{-5}{*}{box\_1}                                & C + G                                                  & 0.045                                                     & 0.070                           & -0.053                          & 0.006                           & 15.71                            & 9.29                             & 27.23                            & 1.26                             & -0.33                            & 0.74                             \\ \hline
                                                        & \cellcolor[HTML]{FFFC9E}GT                             & \cellcolor[HTML]{FFFC9E}0.179                             & \cellcolor[HTML]{FFFC9E}0.090   & \cellcolor[HTML]{FFFC9E}-0.033  & \cellcolor[HTML]{FFFC9E}0.031   & \cellcolor[HTML]{FFFC9E}50.00    & \cellcolor[HTML]{FFFC9E}204.60   & \cellcolor[HTML]{FFFC9E}206.20   & \cellcolor[HTML]{FFFC9E}-54.10   & \cellcolor[HTML]{FFFC9E}51.40    & \cellcolor[HTML]{FFFC9E}-18.80   \\ \cline{2-12} 
                                                        & Baseline-FG                                                 & 0.158                                                     & 0.236                           & 0.341                           & 0.105                           & 2025.66                          & 1088.09                          & 5277.52                          & -1291.44                         & -553.15                          & -271.42                          \\ \cline{2-12} 
                                                        & No C, No G                                             & 0.073                                                     & 0.619                           & 0.352                           & 0.180                           & 1159.95                          & 3027.48                          & 5525.11                          & -1579.93                         & -982.86                          & -555.31                          \\ \cline{2-12} 
                                                        & \cellcolor[HTML]{9AFF99}C, No G                        & \cellcolor[HTML]{9AFF99}0.040                             & \cellcolor[HTML]{9AFF99}0.249   & \cellcolor[HTML]{9AFF99}-0.011  & \cellcolor[HTML]{9AFF99}0.034   & \cellcolor[HTML]{9AFF99}5.49     & \cellcolor[HTML]{9AFF99}234.74   & \cellcolor[HTML]{9AFF99}235.32   & \cellcolor[HTML]{9AFF99}10.12    & \cellcolor[HTML]{9AFF99}-29.32   & \cellcolor[HTML]{9AFF99}1.32     \\ \cline{2-12} 
\multirow{-5}{*}{box\_2}                                & C + G                                                  & 0.039                                                     & 0.253                           & -0.011                          & 0.034                           & 5.56                             & 240.82                           & 241.31                           & 10.24                            & -29.87                           & 1.32                             \\ \hline
                                                        & \cellcolor[HTML]{FFFC9E}GT                             & \cellcolor[HTML]{FFFC9E}0.131                             & \cellcolor[HTML]{FFFC9E}0.099   & \cellcolor[HTML]{FFFC9E}-0.036  & \cellcolor[HTML]{FFFC9E}0.033   & \cellcolor[HTML]{FFFC9E}42.00    & \cellcolor[HTML]{FFFC9E}182.90   & \cellcolor[HTML]{FFFC9E}184.40   & \cellcolor[HTML]{FFFC9E}-45.80   & \cellcolor[HTML]{FFFC9E}44.60    & \cellcolor[HTML]{FFFC9E}-15.50   \\ \cline{2-12} 
                                                        & Baseline-FG                                                 & 0.063                                                     & 0.808                           & 0.304                           & 0.121                           & 707.37                           & 4283.97                          & 5664.51                          & -1598.37                         & -774.69                          & 33.34                            \\ \cline{2-12} 
                                                        & No C, No G                     & 0.061                             & 0.928   & 0.344   & 0.140   & 873.00   & 5442.51  & 7034.36  & -1996.59 & -961.20  & -2.20    \\ \cline{2-12} 
                                                        & \cellcolor[HTML]{9AFF99} C, No G                                                & \cellcolor[HTML]{9AFF99}0.036                                                     & \cellcolor[HTML]{9AFF99}0.180                           & \cellcolor[HTML]{9AFF99}-0.010                          & \cellcolor[HTML]{9AFF99}0.014                           & \cellcolor[HTML]{9AFF99}-0.88                            & \cellcolor[HTML]{9AFF99}127.30                           & \cellcolor[HTML]{9AFF99}125.18                           & \cellcolor[HTML]{9AFF99}5.60                             & \cellcolor[HTML]{9AFF99}-8.83                            & \cellcolor[HTML]{9AFF99}0.38                             \\ \cline{2-12} 
\multirow{-5}{*}{box\_3}                                & C + G                                                  & 0.036                                                     & 0.180                           & -0.010                          & 0.014                           & -0.88                            & 127.29                           & 125.19                           & 5.57                             & -8.79                            & 0.37                             \\ \hline
                                                        & \cellcolor[HTML]{FFFC9E}GT                             & \cellcolor[HTML]{FFFC9E}0.147                             & \cellcolor[HTML]{FFFC9E}0.109   & \cellcolor[HTML]{FFFC9E}-0.035  & \cellcolor[HTML]{FFFC9E}0.031   & \cellcolor[HTML]{FFFC9E}43.30    & \cellcolor[HTML]{FFFC9E}239.00   & \cellcolor[HTML]{FFFC9E}242.00   & \cellcolor[HTML]{FFFC9E}-53.70   & \cellcolor[HTML]{FFFC9E}49.60    & \cellcolor[HTML]{FFFC9E}-16.00   \\ \cline{2-12} 
                                                        & Baseline-FG                                                 & 0.047                                                     & 1.132                           & -0.051                          & 0.289                           & 408.09                           & 6479.25                          & 6157.76                          & 261.21                           & -1542.02                         & 122.95                           \\ \cline{2-12} 
                                                        & No C, No G                                             & 0.045                                                     & 1.062                           & -0.075                          & 0.312                           & 464.67                           & 5512.66                          & 5152.40                          & 354.58                           & -1488.71                         & 115.84                           \\ \cline{2-12} 
                                                        & \cellcolor[HTML]{9AFF99}C, No G                        & \cellcolor[HTML]{9AFF99}0.047                             & \cellcolor[HTML]{9AFF99}0.277   & \cellcolor[HTML]{9AFF99}-0.013  & \cellcolor[HTML]{9AFF99}0.028   & \cellcolor[HTML]{9AFF99}6.27     & \cellcolor[HTML]{9AFF99}328.87   & \cellcolor[HTML]{9AFF99}327.81   & \cellcolor[HTML]{9AFF99}14.49    & \cellcolor[HTML]{9AFF99}-36.91   & \cellcolor[HTML]{9AFF99}1.63     \\ \cline{2-12} 
\multirow{-5}{*}{box\_4}                                & C + G                                                  & 0.047                                                     & 0.277                           & -0.013                          & 0.028                           & 6.20                             & 327.28                           & 326.38                           & 14.38                            & -36.41                           & 1.61                             \\ \hline
\end{tabular}
\end{table}
\begin{table}[h]
  \centering
  \caption{Estimated inertial parameters from our ``In-Hand'' data source while tracking object with ``relaxed-rigidity'' constraints.}
  \label{tab:inertial-est}% Please add the following required packages to your document preamble:
% \usepackage{multirow}
\begin{tabular}{|l|l|r|r|r|r|r|r|r|r|r|r|}
\hline
\multicolumn{1}{|c|}{}                                  & \multicolumn{1}{c|}{}                                  & \multicolumn{1}{c|}{}                                     & \multicolumn{3}{c|}{\textbf{Center of Mass (m)}}                                                    & \multicolumn{6}{c|}{\textbf{Inertia ($10^5$ kg $m^2$) }}                                                                                                                                                         \\ \cline{4-12} 
\multicolumn{1}{|c|}{\multirow{-2}{*}{\textbf{Object}}} & \multicolumn{1}{c|}{\multirow{-2}{*}{\textbf{Method}}} & \multicolumn{1}{c|}{\multirow{-2}{*}{\textbf{mass (kg)}}} & \multicolumn{1}{c|}{\textbf{x}} & \multicolumn{1}{c|}{\textbf{y}} & \multicolumn{1}{c|}{\textbf{z}} & \multicolumn{1}{c|}{\textbf{xx}} & \multicolumn{1}{c|}{\textbf{yy}} & \multicolumn{1}{c|}{\textbf{zz}} & \multicolumn{1}{c|}{\textbf{xy}} & \multicolumn{1}{c|}{\textbf{xz}} & \multicolumn{1}{c|}{\textbf{yz}} \\ \hline
                                                        & \cellcolor[HTML]{FFFC9E}GT                             & \cellcolor[HTML]{FFFC9E}0.115                             & \cellcolor[HTML]{FFFC9E}0.094   & \cellcolor[HTML]{FFFC9E}-0.037  & \cellcolor[HTML]{FFFC9E}0.032   & \cellcolor[HTML]{FFFC9E}37.20    & \cellcolor[HTML]{FFFC9E}150.80   & \cellcolor[HTML]{FFFC9E}154.00   & \cellcolor[HTML]{FFFC9E}-40.00   & \cellcolor[HTML]{FFFC9E}34.80    & \cellcolor[HTML]{FFFC9E}-13.60   \\ \cline{2-12} 
                                                        & Baseline-FG                                                 & 0.048                                                     & 0.385                           & 0.029                           & 0.084                           & 40.46                            & 760.11                           & 781.92                           & -59.37                           & -169.54                          & \cellcolor[HTML]{FFFFFF}14.80    \\ \cline{2-12} 
                                                        & No C, No G                                             & 0.081                                                     & 0.649                           & 0.227                           & 0.090                           & 574.07                           & 3827.66                          & 5589.80                          & -1371.28                         & -866.56                          & 618.89                           \\ \cline{2-12} 
                                                        & \cellcolor[HTML]{9AFF99}C, No G                        & \cellcolor[HTML]{9AFF99}0.044                             & \cellcolor[HTML]{9AFF99}0.249   & \cellcolor[HTML]{9AFF99}-0.012  & \cellcolor[HTML]{9AFF99}0.013   & \cellcolor[HTML]{9AFF99}1.45     & \cellcolor[HTML]{9AFF99}251.28   & \cellcolor[HTML]{9AFF99}258.47   & \cellcolor[HTML]{9AFF99}12.30    & \cellcolor[HTML]{9AFF99}-12.28   & \cellcolor[HTML]{9AFF99}0.35     \\ \cline{2-12} 
\multirow{-5}{*}{box\_0}                                & C + G                                                  & 0.043                                                     & 0.249                           & 0.000                           & 0.011                           & 0.71                             & 237.69                           & 235.80                           & 0.11                             & -17.50                           & 0.04                             \\ \hline
                                                        & \cellcolor[HTML]{FFFC9E}GT                             & \cellcolor[HTML]{FFFC9E}0.179                             & \cellcolor[HTML]{FFFC9E}0.071   & \cellcolor[HTML]{FFFC9E}-0.033  & \cellcolor[HTML]{FFFC9E}0.031   & \cellcolor[HTML]{FFFC9E}50.00    & \cellcolor[HTML]{FFFC9E}165.30   & \cellcolor[HTML]{FFFC9E}166.90   & \cellcolor[HTML]{FFFC9E}-45.10   & \cellcolor[HTML]{FFFC9E}40.70    & \cellcolor[HTML]{FFFC9E}-18.80   \\ \cline{2-12} 
                                                        & Baseline-FG                                                 & 0.135                                                     & 0.386                           & 0.044                           & 0.369                           & 2007.69                          & 5949.03                          & 3166.77                          & 325.31                           & -2325.40                         & -1763.91                         \\ \cline{2-12} 
                                                        & No C, No G                                             & 0.082                                                     & 0.722                           & -0.076                          & 0.311                           & 970.73                           & 6656.01                          & 5165.28                          & 909.11                           & -2171.80                         & -983.14                          \\ \cline{2-12} 
                                                        & \cellcolor[HTML]{9AFF99}C, No G                        & \cellcolor[HTML]{9AFF99}0.048                             & \cellcolor[HTML]{9AFF99}0.180   & \cellcolor[HTML]{9AFF99}-0.018  & \cellcolor[HTML]{9AFF99}0.002   & \cellcolor[HTML]{9AFF99}1.32     & \cellcolor[HTML]{9AFF99}151.14   & \cellcolor[HTML]{9AFF99}152.45   & \cellcolor[HTML]{9AFF99}9.13     & \cellcolor[HTML]{9AFF99}0.46     & \cellcolor[HTML]{9AFF99}-0.03    \\ \cline{2-12} 
\multirow{-5}{*}{box\_1}                                & C + G                                                  & 0.044                                                     & 0.093                           & -0.061                          & 0.000                           & 2.75                             & 6.45                             & 54.04                            & -0.12                            & 1.09                             & -0.19                            \\ \hline
                                                        & \cellcolor[HTML]{FFFC9E}GT                             & \cellcolor[HTML]{FFFC9E}0.179                             & \cellcolor[HTML]{FFFC9E}0.090   & \cellcolor[HTML]{FFFC9E}-0.033  & \cellcolor[HTML]{FFFC9E}0.031   & \cellcolor[HTML]{FFFC9E}50.00    & \cellcolor[HTML]{FFFC9E}204.60   & \cellcolor[HTML]{FFFC9E}206.20   & \cellcolor[HTML]{FFFC9E}-54.10   & \cellcolor[HTML]{FFFC9E}51.40    & \cellcolor[HTML]{FFFC9E}-18.80   \\ \cline{2-12} 
                                                        & Baseline-FG                                                 & 0.204                                                     & 0.166                           & 0.379                           & 0.140                           & 3417.16                          & 968.85                           & 7046.50                          & -1249.28                         & -1057.08                         & -1261.71                         \\ \cline{2-12} 
                                                        & No C, No G                                             & 0.167                                                     & 0.137                           & 0.445                           & 0.127                           & 3926.12                          & 1388.39                          & 6648.07                          & -496.32                          & -1306.25                         & -2505.26                         \\ \cline{2-12} 
                                                        & \cellcolor[HTML]{9AFF99}C, No G                        & \cellcolor[HTML]{9AFF99}0.035                             & \cellcolor[HTML]{9AFF99}0.289   & \cellcolor[HTML]{9AFF99}-0.010  & \cellcolor[HTML]{9AFF99}0.038   & \cellcolor[HTML]{9AFF99}6.24     & \cellcolor[HTML]{9AFF99}264.51   & \cellcolor[HTML]{9AFF99}270.28   & \cellcolor[HTML]{9AFF99}8.83     & \cellcolor[HTML]{9AFF99}-34.90   & \cellcolor[HTML]{9AFF99}1.13     \\ \cline{2-12} 
\multirow{-5}{*}{box\_2}                                & C + G                                                  & 0.035                                                     & 0.302                           & -0.011                          & 0.046                           & 8.42                             & 283.34                           & 288.11                           & 8.15                             & -43.32                           & 1.29                             \\ \hline
                                                        & \cellcolor[HTML]{FFFC9E}GT                             & \cellcolor[HTML]{FFFC9E}0.131                             & \cellcolor[HTML]{FFFC9E}0.099   & \cellcolor[HTML]{FFFC9E}-0.036  & \cellcolor[HTML]{FFFC9E}0.033   & \cellcolor[HTML]{FFFC9E}42.00    & \cellcolor[HTML]{FFFC9E}182.90   & \cellcolor[HTML]{FFFC9E}184.40   & \cellcolor[HTML]{FFFC9E}-45.80   & \cellcolor[HTML]{FFFC9E}44.60    & \cellcolor[HTML]{FFFC9E}-15.50   \\ \cline{2-12} 
                                                        & Baseline-FG                                                 & 0.156                                                     & 0.302                           & 0.528                           & 0.057                           & 4546.66                          & 1693.47                          & 9821.81                          & -2665.26                         & -1027.65                         & 469.28                           \\ \cline{2-12} 
                                                        & No C, No G                     & 0.101                             & 0.504   & 0.534   & 0.068   & 3022.55  & 2767.84  & 8344.82  & -2837.71 & -864.75  & 320.63   \\ \cline{2-12} 
                                                        & \cellcolor[HTML]{9AFF99}C, No G                                                & \cellcolor[HTML]{9AFF99}0.033                                                     & \cellcolor[HTML]{9AFF99}0.241                           & \cellcolor[HTML]{9AFF99}-0.016                          & \cellcolor[HTML]{9AFF99}0.014                           & \cellcolor[HTML]{9AFF99}1.49                             & \cellcolor[HTML]{9AFF99}179.55                           & \cellcolor[HTML]{9AFF99}179.06                           & \cellcolor[HTML]{9AFF99}9.43                             & \cellcolor[HTML]{9AFF99}-13.32                           & \cellcolor[HTML]{9AFF99}0.70                             \\ \cline{2-12} 
\multirow{-5}{*}{box\_3}                                & C + G                                                  & 0.034                                                     & 0.323                           & 0.000                           & 0.030                           & 2.88                             & 307.03                           & 304.18                           & 2.08                             & -26.60                           & 0.18                             \\ \hline
                                                        & \cellcolor[HTML]{FFFC9E}GT                             & \cellcolor[HTML]{FFFC9E}0.147                             & \cellcolor[HTML]{FFFC9E}0.109   & \cellcolor[HTML]{FFFC9E}-0.035  & \cellcolor[HTML]{FFFC9E}0.031   & \cellcolor[HTML]{FFFC9E}43.30    & \cellcolor[HTML]{FFFC9E}239.00   & \cellcolor[HTML]{FFFC9E}242.00   & \cellcolor[HTML]{FFFC9E}-53.70   & \cellcolor[HTML]{FFFC9E}49.60    & \cellcolor[HTML]{FFFC9E}-16.00   \\ \cline{2-12} 
                                                        & Baseline-FG                                                 & 0.223                                                     & 0.258                           & 0.276                           & -0.016                          & 1716.96                          & 2093.23                          & 5416.40                          & -1667.46                         & -64.91                           & 1259.15                          \\ \cline{2-12} 
                                                        & No C, No G                                             & 0.199                                                     & 0.300                           & 0.296                           & -0.025                          & 1760.30                          & 2504.49                          & 5870.91                          & -1840.50                         & 11.30                            & 1431.34                          \\ \cline{2-12} 
                                                        & \cellcolor[HTML]{9AFF99}C, No G                        & \cellcolor[HTML]{9AFF99}0.047                             & \cellcolor[HTML]{9AFF99}0.252   & \cellcolor[HTML]{9AFF99}-0.007  & \cellcolor[HTML]{9AFF99}0.010   & \cellcolor[HTML]{9AFF99}2.82     & \cellcolor[HTML]{9AFF99}284.03   & \cellcolor[HTML]{9AFF99}285.26   & \cellcolor[HTML]{9AFF99}7.91     & \cellcolor[HTML]{9AFF99}-14.98   & \cellcolor[HTML]{9AFF99}0.40     \\ \cline{2-12} 
\multirow{-5}{*}{box\_4}                                & C + G                                                  & 0.045                                                     & 0.247                           & -0.004                          & 0.010                           & 3.37                             & 259.30                           & 261.36                           & 4.62                             & -13.04                           & 0.21                             \\ \hline
\end{tabular}

\end{table}

%%% Local Variables:
%%% mode: latex
%%% TeX-master: "../paperdraft"
%%% End:

\end{appendices}

\end{document}